\documentclass[final]{cvpr}

\usepackage{times}
\usepackage{epsfig}
\usepackage{graphicx}
\usepackage{amsmath}
\usepackage{amssymb}

\usepackage{xcolor}
\usepackage{style}
\usepackage{footnote}
\makesavenoteenv{table}
\makesavenoteenv{tabular}

\usepackage[pagebackref=true,breaklinks=true,colorlinks,bookmarks=false]{hyperref}

\newcommand{\parsection}[1]{\vspace{1mm}\noindent\textbf{#1}~}

\def\cvprPaperID{8054} 
\def\confYear{CVPR 2021}

\begin{document}

\title{Learning Video Instance Segmentation with Recurrent Graph Neural Networks}

\author{Joakim Johnander\\
  Linköping University, Zenseact\\
  {\tt\small joakim.johnander@liu.se}
  \and
  Emil Brissman\\
  Linköping University, SAAB\\
  {\tt\small emil.brissman@liu.se}
  \and
  Martin Danelljan\\
  ETH Zürich\\
  {\tt\small martin.danelljan@vision.ee.ethz.ch}
  \and
  Michael Felsberg\\
  Linköping University\\
  {\tt\small michael.felsberg@liu.se}
}

\maketitle

\begin{abstract}
  Most existing approaches to video instance segmentation comprise multiple modules that are heuristically combined to produce the final output. Formulating a purely learning-based method instead, which models both the temporal aspect as well as a generic track management required to solve the video instance segmentation task, is a highly challenging problem. In this work, we propose a novel learning formulation, where the entire video instance segmentation problem is modelled jointly. We fit a flexible model to our formulation that, with the help of a graph neural network, processes all available new information in each frame. Past information is considered and processed via a recurrent connection. We demonstrate the effectiveness of the proposed approach in comprehensive experiments. Our approach, operating at over 25 FPS, outperforms previous video real-time methods. We further conduct detailed ablative experiments that validate the different aspects of our approach.
  
\end{abstract}

\section{Introduction}

Video instance segmentation is the computer vision task of simultaneously detecting, segmenting, and tracking object instances from a set of predefined classes. This task has a wide range of applications in autonomous driving~\cite{cordts2016cityscapes,yu2020bdd100k}, data annotation~\cite{izquierdo2019prevention,berg2019semi}, and biology~\cite{t2016automatic,zhang2008automatic,burghardt2006analysing}. In contrast to image instance segmentation, the temporal aspect of its video counterpart poses several additional challenges. Preserving correct instance identities in each frame is made difficult by the presence of other, similar instances. Objects may be subject to occlusions, fast motion, or major appearance changes. Moreover, the videos can include wild camera motion and severe background clutter.

Prior works have taken inspiration from the related areas of multiple object tracking, video object detection, instance segmentation, and video object segmentation \cite{yang2019video,athar2020stem,bertasius2020classifying}. Most methods adopt the tracking-by-detection paradigm popular in multiple object tracking \cite{braso2020learning}. An instance segmentation method provides detections in each frame, each detection comprising confidence, semantic class, and mask. The task is then reduced to the formation of \emph{tracks} from these detections. Given a set of already initialized tracks, one must determine for each detection if it belongs to one of the existing tracks, is a false positive, or if it should initialize a new track. Existing approaches learn to match pairs of detections, and then rely on heuristics to form the final output, e.g., initializing new tracks, predicting confidences, removing tracks, and predicting class memberships.

\begin{figure}
  \includegraphics[width=0.48\textwidth, trim={0.0cm 0.0cm 0.0cm 0.0cm}, clip]{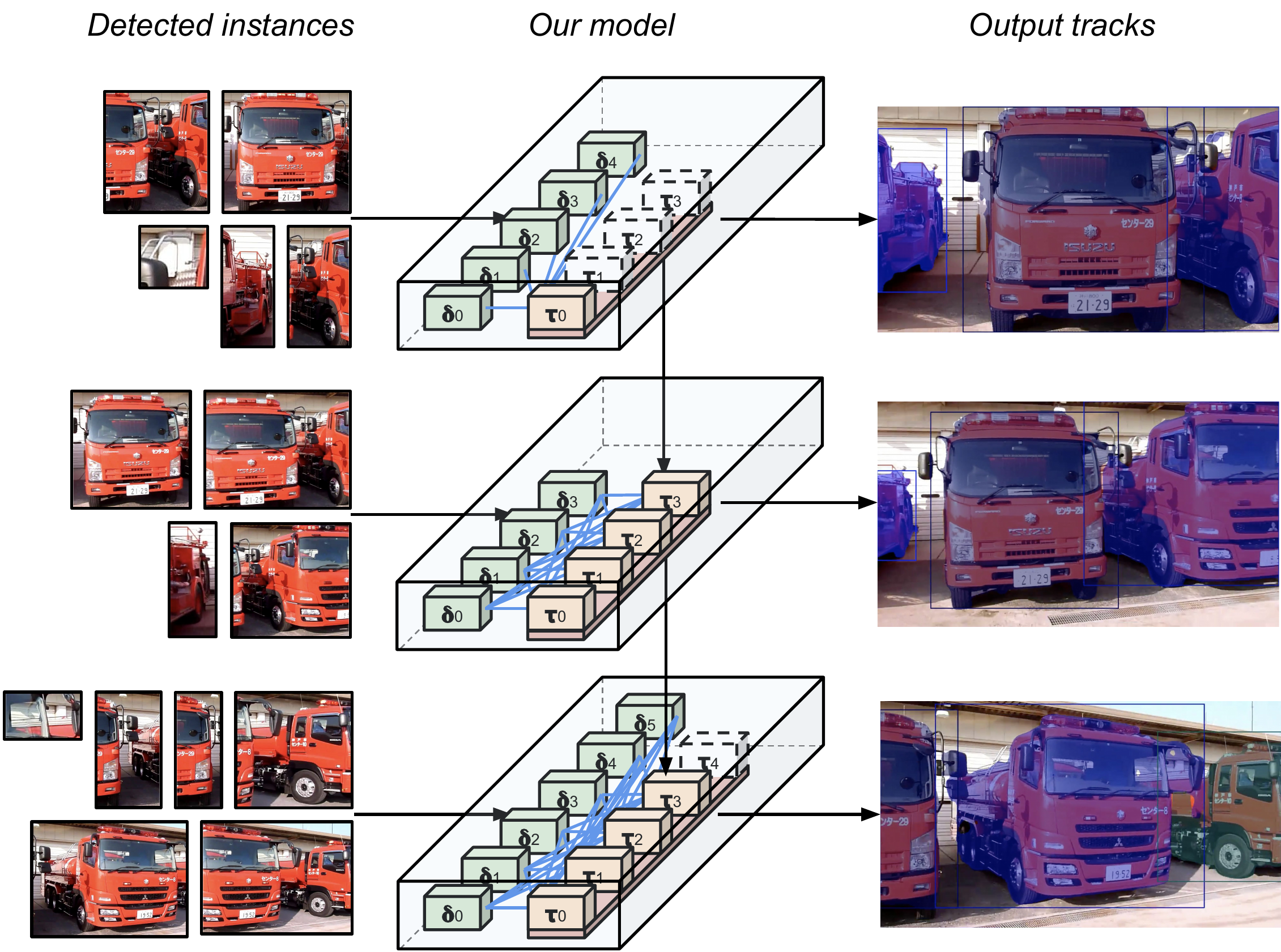}\vspace{1mm}
  \caption{We propose a flexible model that, given detections $\{\delta_n\}_n$ from an image instance segmentation method, builds and updates a memory of tracks $\{\tau_m\}_m$. The model learns to predict probabilities that detections match existing tracks, probabilities that detections should initialize new tracks, and the confidence as well as class membership for each track.}
  \label{fig:illustration}
  \vspace{-6.0mm}
\end{figure}

There are two major drawbacks of such pipelines: (i) the learnt models are not very flexible, for instance not being able to reason globally over all detections or not being able to access information temporally; and (ii) the model learning stage does not closely model the inference, for instance utilizing only pairs of frames or ignoring subsequent detection merging stages. This means that the method does not get the chance to learn the variety of aspects of the video instance segmentation problem. For instance, the method is not able to learn to deal with mistakes made by the utilized instance segmentation method, \eg, due to it being lightweight or the video being challenging.



We address these issues by proposing a novel learning formulation for video instance segmentation that closely models the inference stage during training. Our network proceeds frame by frame, and is in each frame asked to create tracks, associate detections to tracks, and score existing tracks. We use this formulation to train a flexible model, that in each frame processes all available new information via a graph neural network, and considers past information via a recurrent connection. The model predicts, for each detection, a probability that the detection should initialize a new track. The model also predicts, for each pair of an existing track and a detection, the probability that the track and the detection correspond to the same instance. Finally, it predicts an embedding for each existing track that serves two purposes. The embedding is used to predict confidence and class for the track and it is via the recurrent connection fed as input to the GNN in the next frame. 

\newcommand{\contritem}[1]{\textbf{#1}}

\noindent\textbf{Contributions: }
Our main contributions are as follows.
\contritem{(i)} We propose a novel training formulation, allowing us to train models for video instance segmentation in an end-to-end fashion.
\contritem{(ii)} We present a suitable and flexible model based on Graph Neural Networks and Recurrent Neural Networks.
\contritem{(iii)} We model the instance appearance as a Gaussian distribution and introduce a learnable update formulation.
\contritem{(iv)} We benchmark and analyze effectiveness of our approach in comprehensive experiments. Our method outperforms previous near real-time approaches with a relative gain of $9.0\%$ in mAP on the YouTubeVIS dataset~\cite{yang2019video}.

\section{Related Work}
The video instance segmentation (VIS) problem was proposed by Yang et al. \cite{yang2019video} and with it, they proposed several simple and straightforward approaches to tackle the task. They follow the tracking-by-detection paradigm and first apply an instance segmentation method to provide detections in each frame, and then form tracks based on these detections. They experiment with several approaches to matching different detections, such as mask propagation with a video object segmentation method \cite{voigtlaender2019feelvos}; application of a multiple object tracking method \cite{wojke2017simple} in which the image-plane bounding boxes are Kalman filtered and targets are re-detected with a learnt re-identification mechanism; and similarity learning of instance-specific appearance descriptors \cite{yang2019video}. Additionally, they experiment with the offline temporal filtering proposed in \cite{han2016seq}. Cao et al. \cite{cao2020sipmask} propose to improve the underlying instance segmentation method, obtaining both better performance and computational efficiency. Luiten et al. \cite{luiten2019video} propose (i) to improve the instance segmentation method by applying different networks for classification, segmentation, and proposal generation; and (ii) to form tracks with the offline algorithm proposed in \cite{luiten2020unovost}. Bertasius et al. \cite{bertasius2020classifying} also utilize a more powerful instance segmentation method \cite{bertasius2018object}, and propose a novel mask propagation method based on deformable convolutions. Both \cite{luiten2019video} and \cite{bertasius2020classifying} achieve quite remarkable performance, but at a high computational cost.

All of these approaches follow the tracking-by-detection paradigm and try various ways to improve the underlying instance segmentation method or the association of detections. The latter relies mostly on heuristics \cite{yang2019video,luiten2019video} and is often not end-to-end trainable. Furthermore, the track scoring step, where the class and confidence is predicted, has received little attention and is in existing approaches calculated with a majority vote and an averaging operation, as pointed out in the introduction. The work of Athar et al. \cite{athar2020stem} instead propose an end-to-end trainable approach that is trained to predict instance center heatmaps and an embedding for each pixel. A track is constructed from strong responses in the heatmap. The embedding at that location is matched with the embeddings of all other pixels, and if they are sufficiently similar the pixels are assigned to that track.

Our approach is closely related to two works on multiple object tracking (MOT) \cite{braso2020learning,weng2020gnn3dmot} and a work on feature matching \cite{sarlin2020superglue}. These works associate detections or feature points by forming a bipartite graph and applying a Graph Neural Network. The strength of this approach is that the neural network simultaneously reasons about all available information. However, the setting of these works differs significantly from video instance segmentation. MOT is typically restricted to a specific type of scene, such as automotive, and usually with only one or two classes. Furthermore, for both MOT and feature matching, no classification or confidence is to be provided for the tracks. This is reflected in the way \cite{braso2020learning,weng2020gnn3dmot,sarlin2020superglue} utilizes their GNNs, where only either nodes or edges are of interest, not both. The other part exists solely for the purpose of passing messages. As we explain in Sec.~\ref{sec:method}, we will instead utilize both edges and nodes: the edges to predict association and the nodes to predict class membership and confidence.

\section{Method}\label{sec:method}

\begin{figure*}[t]
  \centering
  \includegraphics[width=0.9\textwidth, trim={0.25cm 2.1cm 8.0cm 0.6cm}, clip]{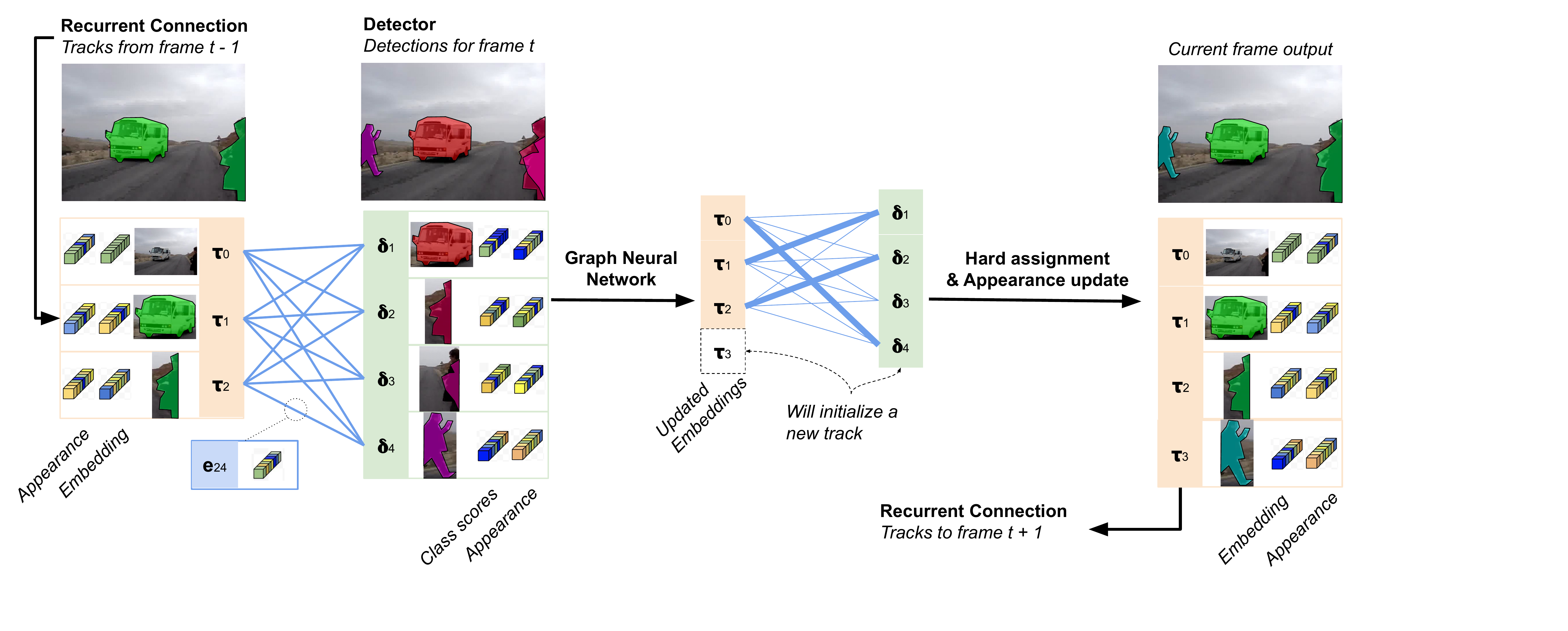}
  \caption{Illustration of the proposed approach. An instance segmentation method is applied to each frame. The set of detections is, together with a maintained memory of tracks, used to construct a graph. Each node and each edge is represented with an embedding. These are processed by a graph neural network and directly used to predict assignment and track initialization (see Section~\ref{sec:gnn}). The track embeddings are further processed by an LSTM-like cell, providing final track embeddings (Section~\ref{sec:rnn}). These are used to predict track confidence and class membership in the current frame (Section~\ref{sec:scoring}), and will via the recurrent connection propagate information to the next time step. Last, tracks are assigned masks and appearance descriptors from matching detections (Section~\ref{sec:appearance}).}
  \label{fig:overview}
\end{figure*}
In this section, we describe our approach in detail. Our aim is to train a model that is able to initialize new tracks, associate detections to existing tracks, and to score existing tracks. Our model works causally, updating tracks in each frame and initializing new ones. First, we use an instance segmentation method to provide tentative detections in each frame. These detections, together with existing tracks, are fed into a graph neural network (GNN). The GNN assesses all available information and provides output embeddings that are used for association and scoring. These output embeddings are furthermore fed as input to the GNN in the next time step, permitting the GNN to process both present and previous information. An overview of the approach is provided in Figure~\ref{fig:overview}.

\subsection{Track-Detection Association}\label{sec:gnn}
In each frame, we let an instance segmentation method produce tentative detections $\{\delta_n\}_n$. Each detection comprises a bounding box, classification scores for each class and the background, and an appearance vector. We maintain a memory $\{\tau_m\}_m$ of previously seen objects, usually referred to as tracks. The aim of our model is to associate existing tracks with the detections, determining whether or not a track $\tau_m$ corresponds to a detection $\delta_n$. Additionally, the model needs to decide whether a detection $\delta_n$ should initialize a new track. Note that the detections $\{\delta_n\}_n$ are in general noisy, and correctly deciding whether a detection should initialize a new track may be outright impossible. In such scenarios we expect the model to create a track, and over time as more information is accumulated, re-evaluate whether the track is a true positive or not.

Most existing methods \cite{yang2019video,luiten2019video,cao2020sipmask} associate tracks to detections by training a network to extract appearance descriptors. The descriptors are trained to be similar if they correspond to the same object, and dissimilar if they correspond to different objects. The issue with such an approach is that appearance descriptors corresponding to visually and semantically similar, but different instances, will be trained to be different. In such scenarios it might be better to let the appearance descriptors be similar, and instead rely on for instance spatial information. The network should therefore assess all available information before making its decision.

Further information is obtained from track-detection pairs other than the one considered. It may be difficult to determine whether a track-detection pair matches in isolation, for instance with cluttered scenes or when visibility is poor. In such scenarios, the instance segmentation method might provide multiple detections that all overlap the same object to some extent. Another difficult scenario is when there is sudden and severe camera motion, in which case we might need global reasoning in order to either disregard spatial similarity or treat it differently. We therefore hypothesize that it is important for the network to reason about all tracks and detections simultaneously.

The same is naturally true when determining whether a detection should initialize a new track. How well the detection matches existing tracks must influence the decision on whether it should initialize a new one. In existing work \cite{yang2019video,cao2020sipmask}, this observation is implemented as a hard decision. That is, if the detection is assigned to a track it will not initialize a new one, otherwise it will. We avoid this heuristic and instead let the network simultaneously reason about assigning detections to tracks and initializing new tracks.

Let each track-detection pair be represented with a $D$-dimensional embedding $e_{mn}\in\mathbb{R}^D$. Initially, it contains information on the similarity or relationship between track $\tau_m$ and detection $\delta_n$. Further, let each potential new track be represented with an embedding $e_{0n}\in\mathbb{R}^D$. We create an empty track embedding $\tau_0$, and initialize $e_{0n}$ with information on the detection $\delta_n$ as well as its similarity to the empty track $\tau_0$. We treat the empty track the way we treat other tracks, but it is processed with its own set of weights. We illustrate the elements $\tau_m$, $\delta_n$, and $e_{mn}$ in Figure~\ref{fig:overview}. The detection embeddings $\delta_n$ are initialized with the class scores provided by the detector, as well as the bounding boxes. The track embeddings $\tau_m$ are the final track embeddings output in the previous time step. We now propagate information between the different embeddings in a learnable way. To this end we use layers that perform updates of the form
\begin{subequations}
  \label{eq:gnn}
  \begin{align}
    e_{mn}^+ &= f^e([e_{mn}, \tau_m, \delta_n]) \enspace,\label{eq:gnn1}\\
    \tau_m^+ &= f^\tau([\tau_m, \sum_j g^\tau(e_{mj})e_{mj}]) \enspace,\label{eq:gnn3}\\
    \delta_n^+ &= f^\delta([\delta_n, \sum_i g^\delta(e_{in})e_{in}]) \enspace.\label{eq:gnn4}
  \end{align}
\end{subequations}
Here, $f^e$, $f^\tau$, and $f^\delta$ are linear layers followed by a ReLU activation, $g^\tau$ and $g^\delta$ are multilayer perceptrons ending with the logistic function, and $[\cdot,\cdot]$ denotes concatenation. This type of layer has the structure of a Graph Neural Network (GNN) block \cite{DBLP:journals/corr/abs-1806-01261}, with both $\tau_m$ and $\delta_n$ as nodes, and $e_{mn}$ as edges. These layers permit information exchange between the embeddings. The layer deviates slightly from the literature. First, we have two types of nodes and use two different updates for them. This is similar to the work of Brasó et al. \cite{braso2020learning} where message passing forward and backward in time uses two different neural networks. Second, the accumulation in the nodes in \eqref{eq:gnn3} and \eqref{eq:gnn4} uses an additional gate, permitting the nodes to dynamically select from which message information should be accumulated. This is sensible in our setting, as for instance class information should be passed from detection to track if and only if the track and detection match well. We construct our graph neural network by stacking GNN blocks. For added expressivity, we interleave them with standard residual blocks, in which there is no information exchange between different graph elements. The GNN will provide us with updated edge embeddings which we use for association of detections to tracks, and updated node embeddings which will be used to score tracks and as input to the GNN in the next frame.


The edge embeddings $e_{mn}$ are fed through a logistic model to predict the probability that the track $m$ matches the detection $n$. If the probability is high, they are considered to match and the track will obtain the segmentation of that detection. New tracks are initialized in a similar fashion. The edge embeddings $e_{0n}$ are fed through another logistic model to predict the probability that the detection $n$ should initialize a new track. If the probability is beyond a threshold, we initialize a new track with the embedding of that detection $\delta_n$. This threshold is intentionally selected to be quite low, as we always have the possibility to lower the score as we progress through time. Next, we describe how tracks are scored.

\subsection{Track Scoring}\label{sec:scoring}
For each created track, a confidence and a class prediction are reported. The confidence reflects our trust about whether the track is a true positive and it is updated over time together with the class prediction as more information becomes available. This provides the model with the option of effectively removing tracks by reducing their scores. Existing approaches \cite{cao2020sipmask,yang2019video,luiten2019video,bertasius2020classifying} score tracks by averaging the detection confidence of the detections deemed to correspond to the track. Class predictions are made with a majority vote. The drawback is that other available information, such as how certain we are that each detection indeed belongs to the track or the consistency of the detections, is not taken into account. This issue is addressed by the GNN introduced in Sec.~\ref{sec:gnn} together with the recurrent connection introduced in Sec.~\ref{sec:rnn}. The track embeddings $\{\tau_m\}_m$ gather information from all detections produced that frame via the GNN, and accumulate this information over time via the recurrent connection. We then predict the confidence and class for each track based on its embedding via a multinomial logistic model.

We furthermore report a segmentation for each track. Each track assumes the segmentation provided by the detection with which it is considered to match. If a track matches multiple detections, it selects the best match. Furthermore, we report only a single instance for each pixel. In the presence of false positive tracks, we therefore need to reweight the segmentations of the different tracks. To this end, we concatenate the segmentation and the embedding of the corresponding track and feed it through a two-layer convolutional neural network.


\subsection{Recurrent Connection}\label{sec:rnn}
In order to process object tracks, it is crucial to propagate information over time. We achieve this with a recurrent connection, which brings the benefit of end-to-end training. However, naïvely adding recurrent connections leads to highly unstable training and in extension, poor video instance segmentation results. Even with careful weight initialization and low learning rate, both activation and gradient spikes arise.

This is a well-known problem when training recurrent neural networks and is usually tackled with the Long Short-Term Memory (LSTM) \cite{hochreiter1997lstm} or Gated Recurrent Unit (GRU) \cite{cho2014properties} modules. These modules use a system of multiplicative sigmoid-activated \emph{gates}, and have been repeatedly shown to be able to well model sequential data \cite{hochreiter1997lstm,cho2014properties,greff2016lstm}. The vanilla LSTM has the form
\begin{subequations}
  \label{eq:rnn}
  \begin{align}
    z_i^t &= [x_i^t, y_i^{t-1}]\enspace,\\
    \alpha_i^{\text{forget}} &= \sigma(h^{\text{forget}}(z_i^t))\enspace,\label{eq:rnn:forget}\\
    \alpha_i^{\text{input}}  &= \sigma(h^{\text{input}} (z_i^t))\enspace,\\
    \alpha_i^{\text{output}} &= \sigma(h^{\text{output}}(z_i^t))\enspace,\\
    \tilde{c}_i^t &= \text{tanh}(h^{\text{cell}}(z_i^t))\enspace,\\
    c_i^t &= \alpha_i^{\text{forget}}\odot c_i^{t-1} + \alpha_i^{\text{input}}\odot \tilde{c}_i^t\enspace,\\
    y_i^t &= \alpha_i^{\text{output}}\odot \text{tanh}(c_i^t)\enspace,\label{eq:rnn:y}
  \end{align}
\end{subequations}
where $x_i^t$, $y_i^t$, and $c_i^t$ are the input, output, and so-called cell-state at time step $t$. $h^{\text{forget}}, h^{\text{input}}, h^{\text{output}}, h^{\text{cell}}$ are linear neural network layers. $\odot$ is the element-wise product, $\text{tanh}$ the hyperbolic tagent, and $\sigma$ the logistic function.

The recurrent connection in the LSTM, where its output is fed as input to the next time step, allows it to model temporal information. Its system of gates alleviates problems with exploding and vanishing gradients and activations.

However, our scenario differs slightly from those typical to the LSTM. We want to recurrently feed output of our graph network, specifically the updated track embeddings, as input to the graph network in the next time step. We hypothesize that we can replace the linear layer in the LSTM with our entire graph network and still benefit from the gating mechanism. We let equations \eqref{eq:rnn:forget} - \eqref{eq:rnn:y} establish a mapping $L:z\rightarrow y$, and update the track embeddings as
\begin{subequations}
  \label{eq:rnngnn}
  \begin{align}
    \{\tilde{\tau}_m^t\}_m &= \text{GNN}(\{\tau_m^{t-1}\}_m, \{\delta_n^t\}_n)\enspace,\\
    \tau_m^t &= L(\tilde{\tau}_m^t),\quad\forall m\enspace.
  \end{align}
\end{subequations}
The track embeddings output by the GNN $\{\tilde{\tau}_m^t\}_m$ are fed into the system of gates and the output is used as our final track embeddings $\{\tau_m^t\}_m$. These are used both as input to our GNN in the next time step and to score the tracks.

\subsection{Modelling Appearance}\label{sec:appearance}
In order to accurately match tracks and detections, we create instance-specific appearance models for each tracked object. To this end, we add an appearance network, comprising a few convolutional layers, and apply it to feature maps of the backbone ResNet \cite{he2016deep}. The output of the appearance network is pooled with the masks provided by the detections, resulting in an appearance descriptor for each detection. The tracks gather appearance from the detections and over time construct an appearance model of that track. The similarity in appearance between a track and a detection will serve as an important additional cue during matching. The aim for the appearance network is to learn a rich representation that allows us to discriminate between visually or semantically similar instances.


Our initial experiments of integrating appearance information directly into the GNN, similar to \cite{braso2020learning,sarlin2020superglue,weng2020gnn3dmot}, did not lead to noticeable improvement. This is likely due to differences between the problems. The video instance segmentation problem is fairly unconstrained: there is significant variation in scenes and objects considered, and compared to its variation, there are quite few labelled training sequences available. In contrast, multiple object tracking typically works with a single type of scene or a single category of objects and feature matching is learnt with magnitudes more training examples than what is available for video instance segmentation.

In order to sidestep this issue, we treat appearance separately and enforce symmetry across the feature channels. Each track models its appearance as a multidimensional Gaussian distribution with diagonal covariance. When the track is initialized, we use the appearance vector of the initializing detection as mean $\mu$ and a fixed covariance $\Sigma$. We feed appearance information into the GNN via the track-detection edges. The edge between track $m$ and detection $n$ is initialized with the loglikelihood of the detection appearance given the track distribution. The GNN is able to utilize this information when calculating the matching probability of each track-detection pair. Afterwards, the appearance $(\mu,\Sigma)$ of each track is updated with the appearance $x$ of the best matching detection. The update is based on the Bayesian update of a Gaussian under a conjugate prior. We use a normal-inverse-chi-square prior \cite{murphy2007conjugate},
\begin{subequations}
  \begin{align}
    \mu^+ &= \kappa x + (1 - \kappa)\mu\enspace,\label{eq:app:mean}\\
    \Sigma^+ &= \nu\tilde{\Sigma} + (1 - \nu)\Sigma + \frac{\kappa(1 - \nu)}{\kappa + \nu}(x - \mu)^2\enspace.\label{eq:app:cov}
  \end{align}
\end{subequations}
The term $\tilde{\Sigma}$ corresponds to the sample variance and the update rates $\kappa$ and $\nu$ would usually be the number of samples in the update relative the strength of the prior. For added flexibility we predict these values based on the track embedding, permitting the network to learn a good update strategy.

\subsection{Training Procedure}
We train the network by feeding a sequence of $T$ frames through it as we would do during inference at test-time. The network provides a set of tracks, each containing per-frame scores for each class and the background, boxes, and segmentations. We apply four losses,
\begin{align}
  \mathcal{L} = \lambda^1\mathcal{L}^\text{score} + \lambda^2\mathcal{L}^\text{seg} + \lambda^3\mathcal{L}^\text{match} + \lambda^4\mathcal{L}^\text{init}\enspace.
  \label{eq:loss}
\end{align}
The component $\mathcal{L}^\text{score}$ rewards the network for predicting the correct class scores; $\mathcal{L}^\text{seg}$ for correctly reweighting segmentations; $\mathcal{L}^\text{match}$ for correctly assigning detections to each track after initialization; and $\mathcal{L}^\text{init}$ for correctly initializing new tracks.

The first step towards constructing these loss functions is to determine for each detection whether it matches an annotated object and if so, which one. We say that a detection matches an annotated object if their bounding boxes overlap with at least $0.5$ intersection over union \cite{everingham2015pascal}. If multiple detections match the same annotated object, we assign the annotation to the best matching detection and mark the remainder as negatives. This procedure is repeated for each frame, after which all detections have been marked as background or with an annotation identity. Next, we assign an identity to each track. During inference, we mark the detection that initialized the track. We then give the track the same annotation identity as the detection that initialized it.

We are now ready to define the losses. For each track, we check whether it is marked as background or the class of its annotation identity and apply a cross-entropy loss $\mathcal{L}^\text{score}$. We noted that it is difficult to score tracks early on in some scenarios and therefore weight the normalization over the sequence, giving a higher weight to later frames. Next, we map the indices of the annotated instance segmentation to our track identities and apply the Lovasz loss \cite{berman2018lovasz}. This forces the network to deal with cases where multiple tracks claim the same pixels by reweighting the segmentation scores. We apply binary cross-entropy losses $\mathcal{L}^\text{match}$ and $\mathcal{L}^\text{init}$ to predicted track-detection matching probabilities and track initialization probabilities. We say that a detection should initialize a new track if the detection was matched to an annotation for which there is no track yet. We normalize these losses with the batchsize and sequence length, but not with the number or tracks or detections. Detections or tracks that are false positives will therefore not reduce the loss for other tracks or detections, as they otherwise would.
\subsection{Implementation Details}
We implemented the proposed approach in PyTorch \cite{pytorch} and will make code available upon publication. Training and inference is done with a single Nvidia V100 GPU. We held out 200 sequences of the YouTubeVIS \cite{yang2019video} training set for hyperparameter selection. We use the publicly available implementation of the instance segmentation method YOLACT \cite{bolya2019yolact} with a ResNet50 \cite{he2016deep} backbone. The backbone and detector are initialized using the weights supplied with the implementation. The class-dependent detector layers are replaced to output the number of classes in YouTubeVIS. We finetune the backbone and detector for 120 epochs á 933 iterations using images from YouTubeVIS \cite{yang2019video} and OpenImages \cite{OpenImages,OpenImagesSegmentation}. We adopt random scaling, crops, and horizontal flips. We optimize with Adam, using a batch size of 8, a learning rate of $5\cdot 10^{-5}$ that is decayed by a factor of 5 after epochs 60 and 90, and a weight decay of $10^{-4}$. Next, we freeze the backbone and the detector, and train all other modules, the appearance network, the GNN, and the recurrent gating module. We train for 150 epochs á 633 iterations with video clips of 10 frames sampled randomly from YouTubeVIS. We again adopt random scaling, crops, and  horizontal flips. We optimize with Adam, using a batch size of 4, a learning rate of $2\cdot 10^{-4}$, and a weight decay of $10^{-4}$. The loss weights in \eqref{eq:loss} are set to $1$, $1$, $4$, and $1$ respectively. The initial covariance for the track appearance distribution is set to $0.001$. We use two GNN blocks interleaved with two residual blocks and $D=128$ embedding channels. We set the track initialization threshold probability to $0.31$ ($\text{softplus}(-1)$) during training and $0.13$ during inference. Tracks are marked as inactive if there is no detection that matches with a probability of at least $0.31$.

\section{Experiments}\label{sec:experiments}
\begin{figure*}[t]
  \vspace{-1.5mm}
  \hspace{18.5mm}
  \includegraphics[width=0.22\textwidth,trim={1.0cm 0.5cm 1.0cm 0.5cm}, clip]{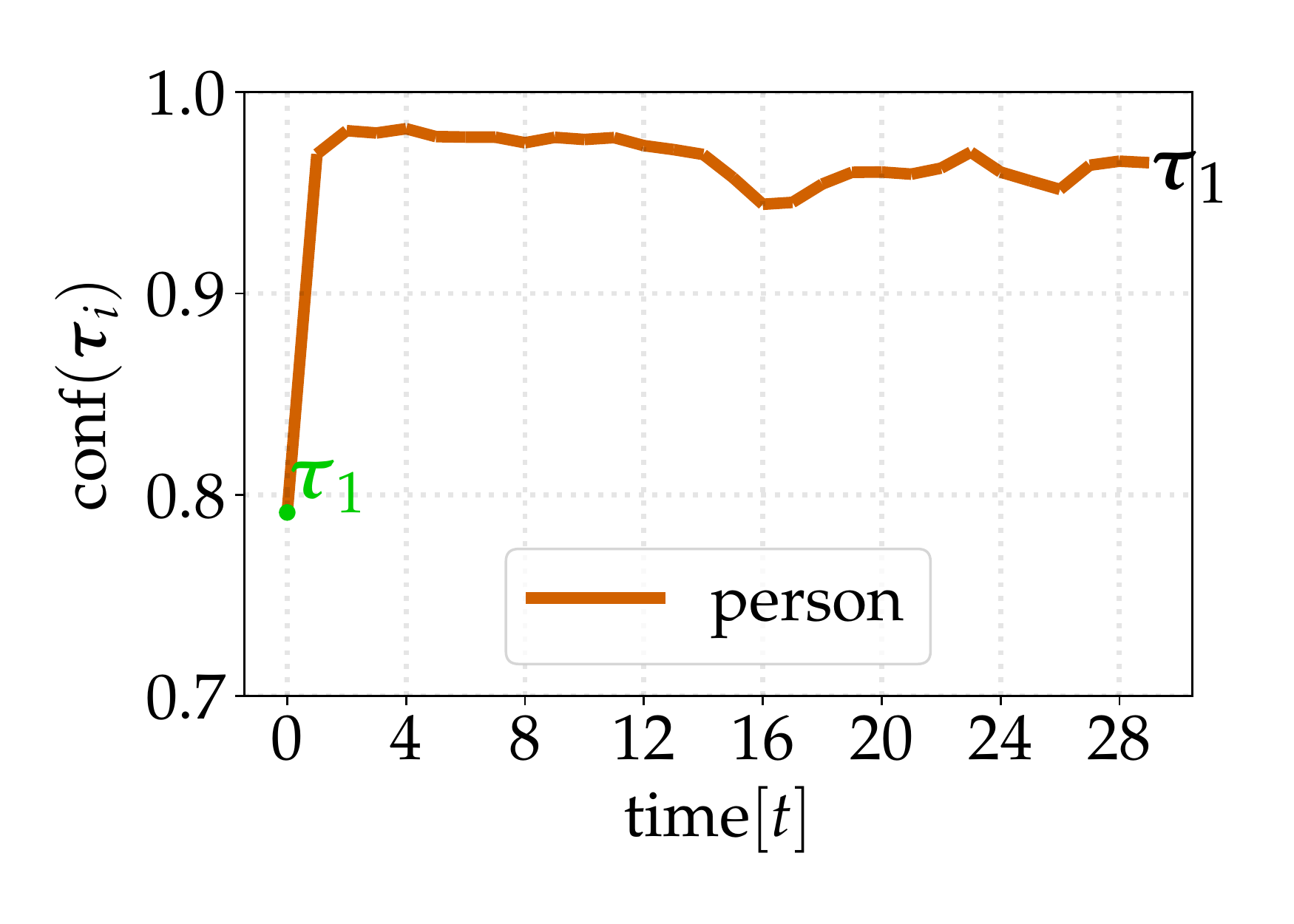} %
  \hspace{17.5mm}
  \includegraphics[width=0.22\textwidth,trim={1.0cm 0.5cm 1.0cm 0.5cm}, clip]{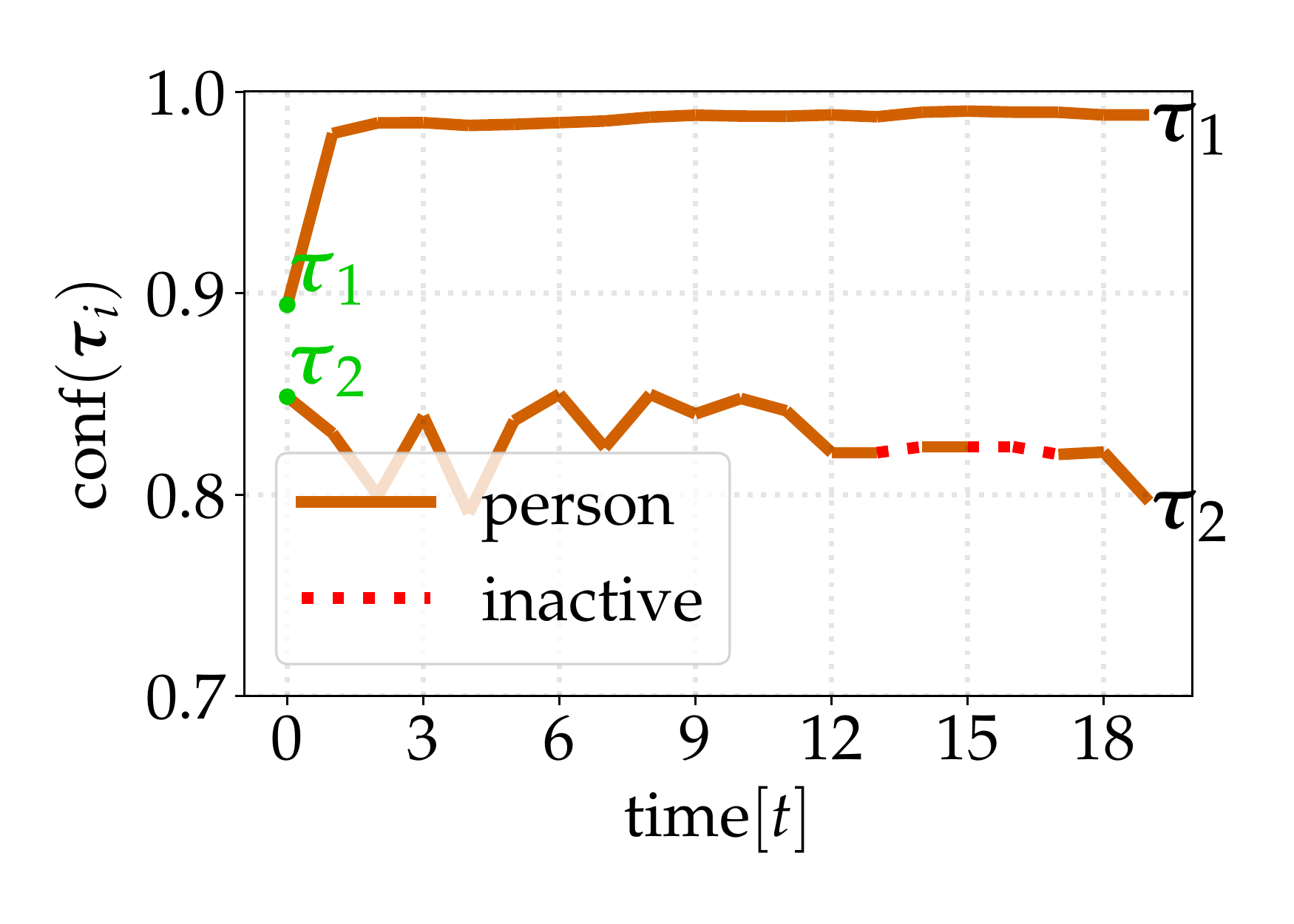} %
  \hspace{18.0mm}
  \includegraphics[width=0.215\textwidth,trim={1.0cm 0.5cm 1.0cm 0.5cm}, clip]{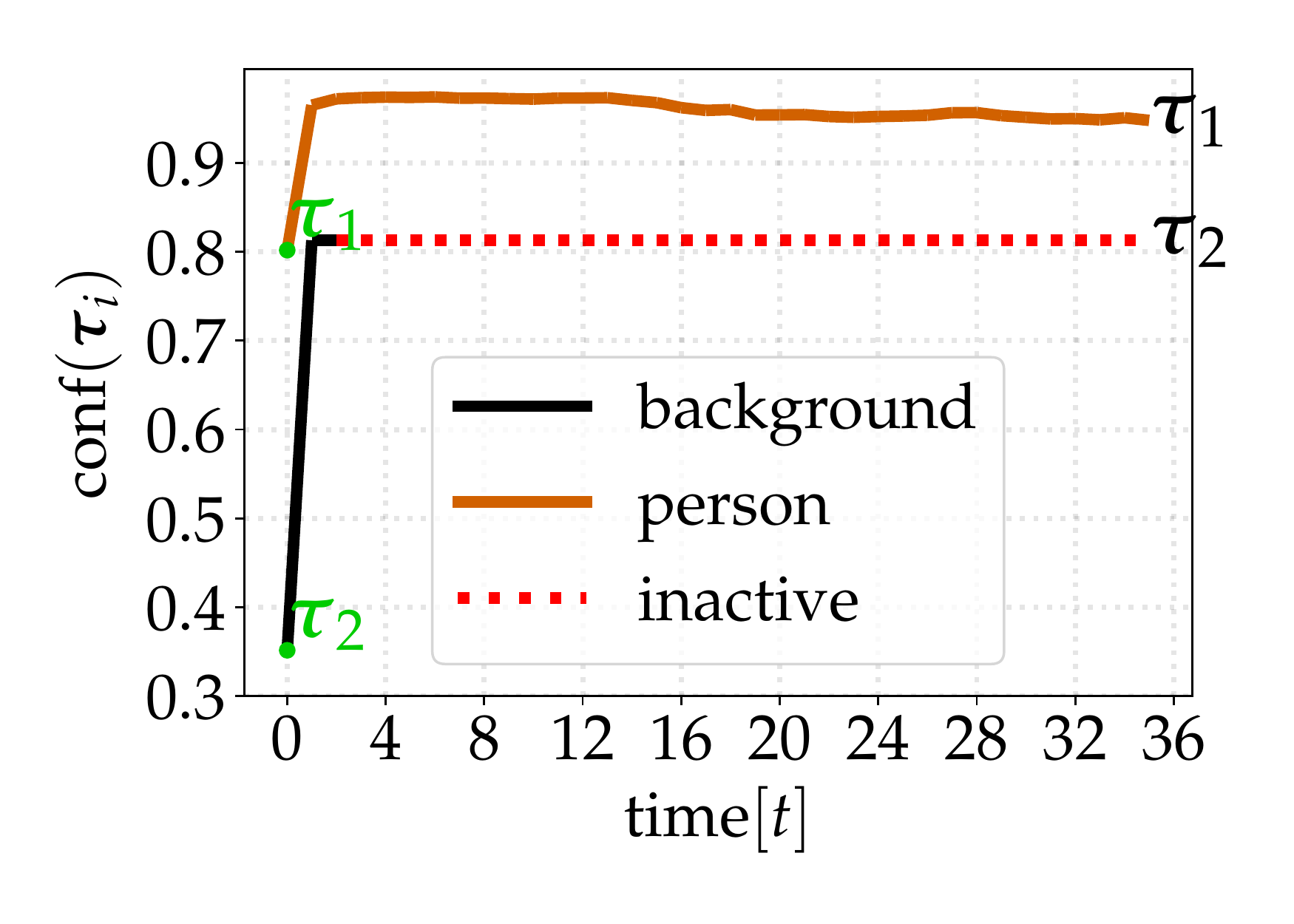} \\
  \centering
  \includegraphics[width=0.99\textwidth,trim={0.0cm 0.0cm 0.0cm 0.0cm}, clip]{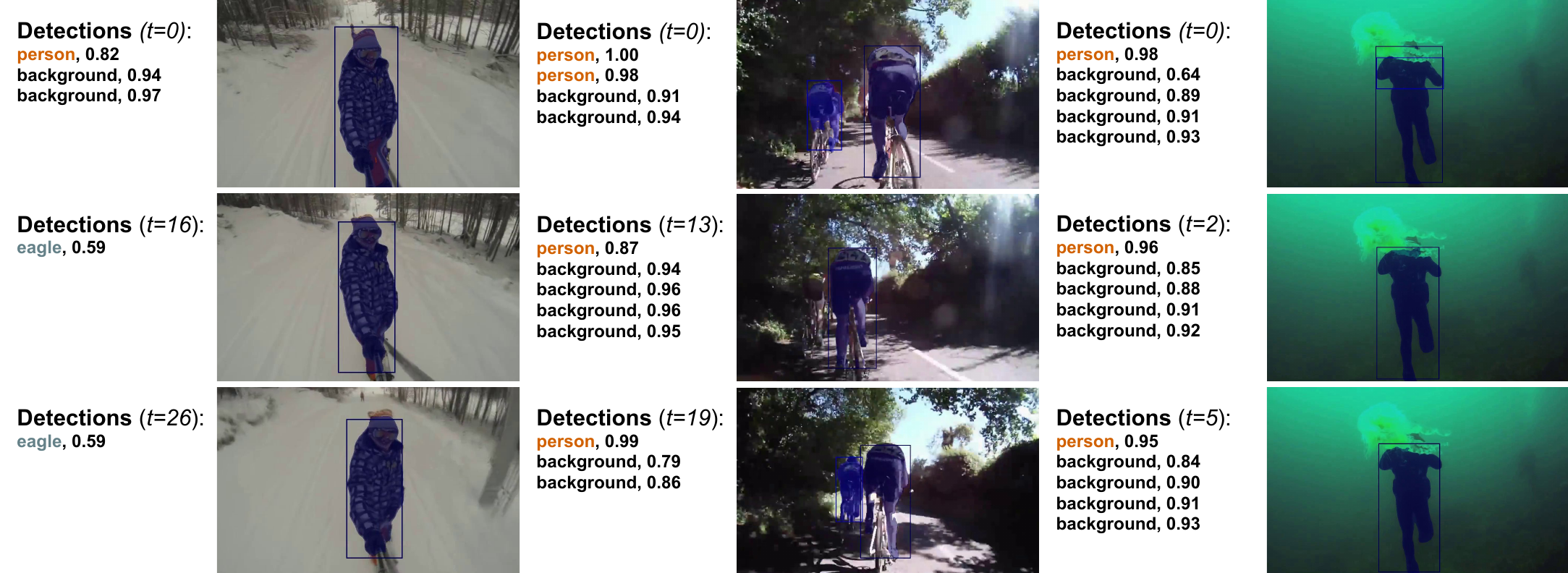}
  

  \caption{Columnwise list of detections and track images for three videos. The plots on the top show ground-truth classes and the corresponding predicted probabilities. Note that \emph{background} is included in the classes, and higher is therefore always better. In the left column, the detector makes noisy class predictions, but our approach is able to filter this noise. In the center column there is a missed detection for the left cyclist. Our method does as well as it can, rendering that track inactive and resumes it in subsequent frames when the detector again finds both objects. To the right, a false positive in the detector leads to a false track. This track is however quickly marked with low confidence (marked as the \emph{background class}). These examples show that the approach have learnt to suppress noise from the detector.}
  \label{fig:exp_3in1}
  \vspace{-3mm}
\end{figure*}

We evaluate the proposed approach for video instance segmentation on YouTubeVIS~\cite{yang2019video}, a benchmark comprising 40 object categories in 2883 videos. Performance is measured in terms of video mean average precision (mAP). We first conduct an ablation study to analyze different aspects of our approach, we then compare to the state-of-the-art, and last we qualitatively assess performance.


\subsection{Ablation Study}
In our ablative experiments we make five key modifications to our approach, which are described below. Results are shown in Table~\ref{tab:ablation}. For each study we use the same hyperparameter setup and keep the detector fixed.

\begin{table}[t]
  \centering%
  \resizebox{0.7\columnwidth}{!}{%
  \begin{tabular}{l c}\hline
    \textbf{Configuration} & \textbf{mAP}\\\hline
    Proposed approach & 35.3\\\hline
    Limited GNN & 28.6\\
    Association from \cite{yang2019video} & 29.2\\\hline
    No appearance & 34.5\\
    Appearance const. variance & 33.0\\
    Appearance update with LSTM & 34.0\\
    Appearance baked into embedding & 34.4\\\hline
    Scoring from \cite{yang2019video} & 31.5\\
    Scoring via yolact average & 30.7\\\hline
    No LSTM-like gating & Diverges\\
    Simple gate & 31.5\\\hline
    ResNet101 & 37.7\\\hline
  \end{tabular}}
  \vspace{1mm}
  \caption{Performance under different configurations on the YouTubeVIS validation set.}%
  \label{tab:ablation}%
  \vspace{-4mm}
\end{table}

\parsection{Limited GNN} We argue in Sec. \ref{sec:gnn} that it is important to process all tracks and detections simultaneously. We verify this statement by limiting the GNN to the greatest possible extent. For track-detection assignment, we remove the information flow in the graph and rely solely on the appearance and spatial similarity of that pair. In order to update tracks, they need to gather information from detections. We therefore let each track gather information once from its best matching detection. New tracks are initialized with a hard decision similar to~\cite{yang2019video}. With the limited GNN, we observe a 6.7 absolute drop in mAP.

\begin{table}[t]
  \vspace{2mm}%
  \centering%
  \resizebox{0.65\columnwidth}{!}{%
  \begin{tabular}{l c c}\hline
    \textbf{Method}                          & \textbf{fps} & \textbf{mAP}\\\hline
    OSMN MaskProp \cite{OSMN}                &    & 23.4\\
    FEELVOS \cite{voigtlaender2019feelvos}   &    & 26.9\\
    IoUTracker+ \cite{yang2019video}         &    & 23.6\\
    OSMN \cite{OSMN}                         &    & 27.5\\
    DeepSORT \cite{wojke2017simple}          &    & 26.1\\
    SeqTracker \cite{yang2019video}          &    & 27.5\\
    MaskTrack R-CNN \cite{yang2019video}     & 20 & 30.3\\
    SipMask \cite{cao2020sipmask}            & 30 & 32.5\\
    SipMask ms-train \cite{cao2020sipmask}   & 30 & 33.7\\
    STEm-Seg ResNet50 \cite{athar2020stem}   &    & 30.6\\
    STEm-Seg ResNet101 \cite{athar2020stem}  &  7 & 34.6\\
    VIS2019 Winner \cite{luiten2019video}    & $<1^\dagger$ & 44.8\\
    MaskProp \cite{bertasius2020classifying} & $<2^\ddagger$ & 46.6\\\hline
    \textbf{Ours} (ResNet50)                 & 30 & \second{35.3}\\
    \textbf{Ours} (ResNet101)                & 25 & \first{37.7}\\\hline
  \end{tabular}}
  \vspace{1mm}
  \caption{State-of-the-art comparison on the YouTubeVIS validation dataset~\cite{yang2019video}. The proposed approach outperforms all near real-time approaches. $\dagger$ and $\ddagger$: No speed reported in \cite{luiten2019video} or \cite{bertasius2020classifying}, but each utilize components (\cite{luiten2020unovost} and \cite{chen2019hybrid}) which are reported with a speed of 1 fps and 2 fps respectively.}\vspace{-3mm}
  \label{tab:quantsota}
\end{table}

\parsection{Association from \cite{yang2019video}} We compare with a simpler method for association. Instead of letting the GNN predict the association probability, we utilize the rule from \cite{yang2019video}. The association score is calculated as a linear combination of the appearance similarity, intersection over union, Kronecker delta of the class, and detection confidence. This leads to a 6.1 absolute drop in mAP.

\parsection{Appearance} We measure the importance of the appearance by removing it. We experiment with a constant covariance. We furthermore experiment with letting an LSTM directly predict mean and variance. Last, we remove the separate treatment of appearance and bake it into the embeddings. All these configurations lead to deteriorated performance.

\begin{figure*}[t]
  \centering
  \includegraphics[width=0.25\textwidth,trim={3.0cm 2.0cm 3.5cm 2.0cm}, clip]{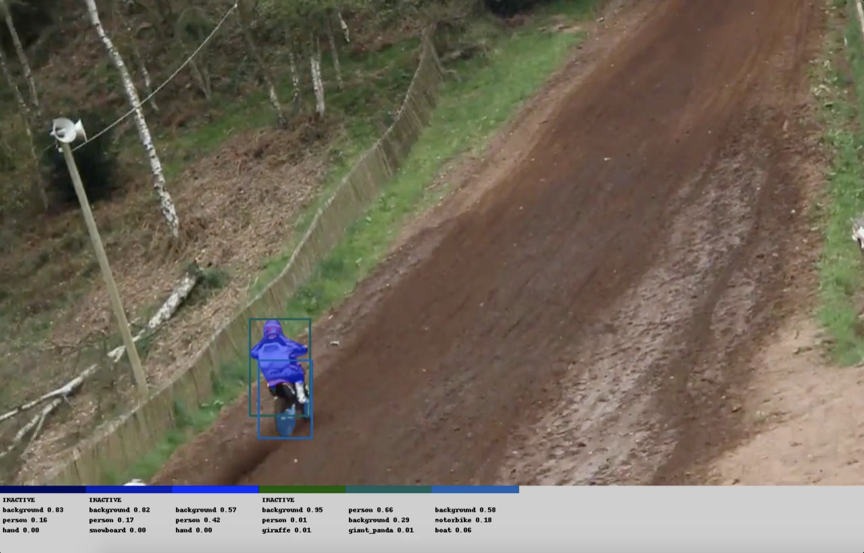}%
  \includegraphics[width=0.25\textwidth,trim={3.0cm 2.0cm 3.5cm 2.0cm}, clip]{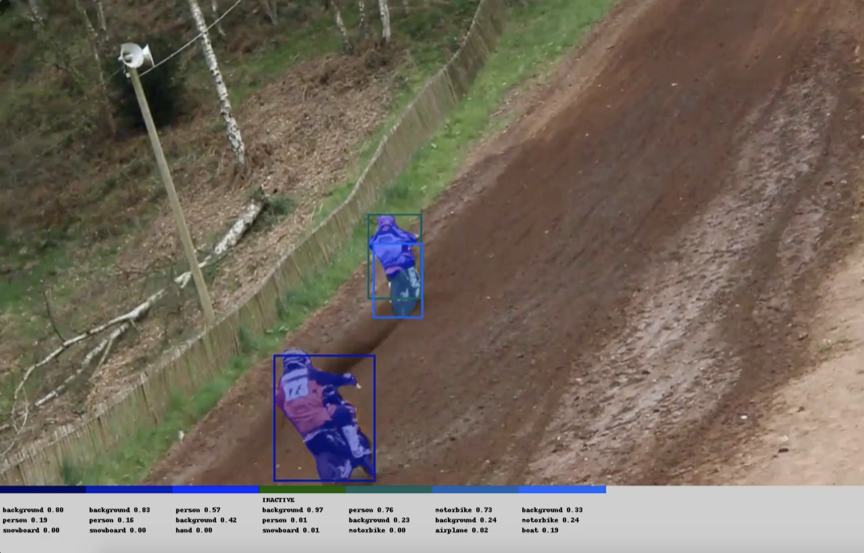}%
  \includegraphics[width=0.25\textwidth,trim={3.0cm 2.0cm 3.5cm 2.0cm}, clip]{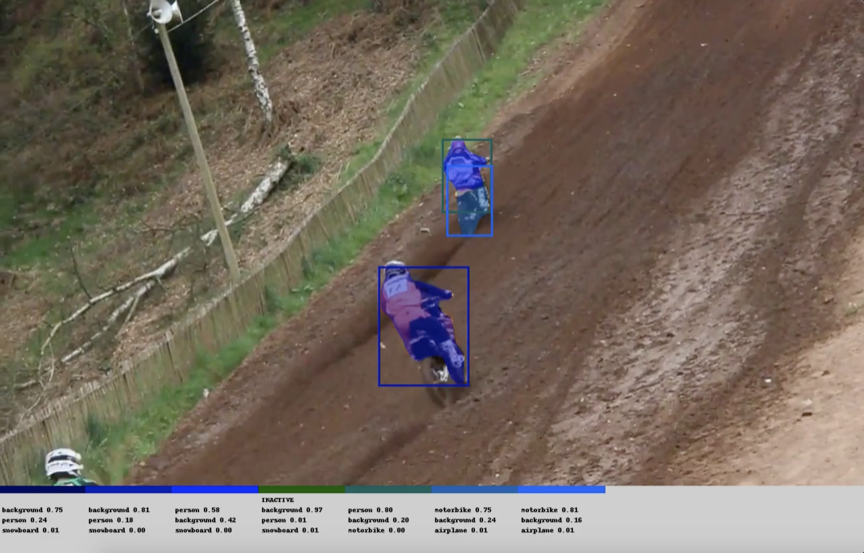}%
  \includegraphics[width=0.25\textwidth,trim={3.0cm 2.0cm 3.5cm 2.0cm}, clip]{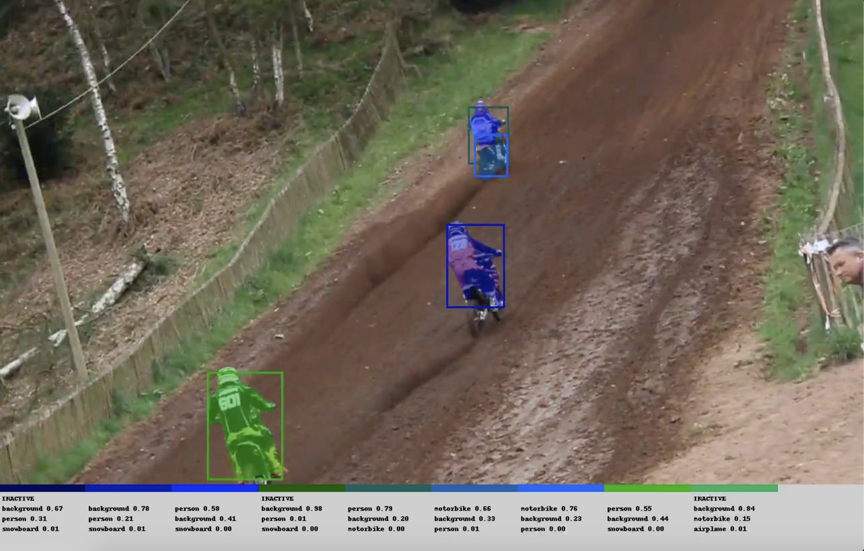}
  \includegraphics[width=0.25\textwidth,trim={5.5cm 2.0cm 1.0cm 2.0cm}, clip]{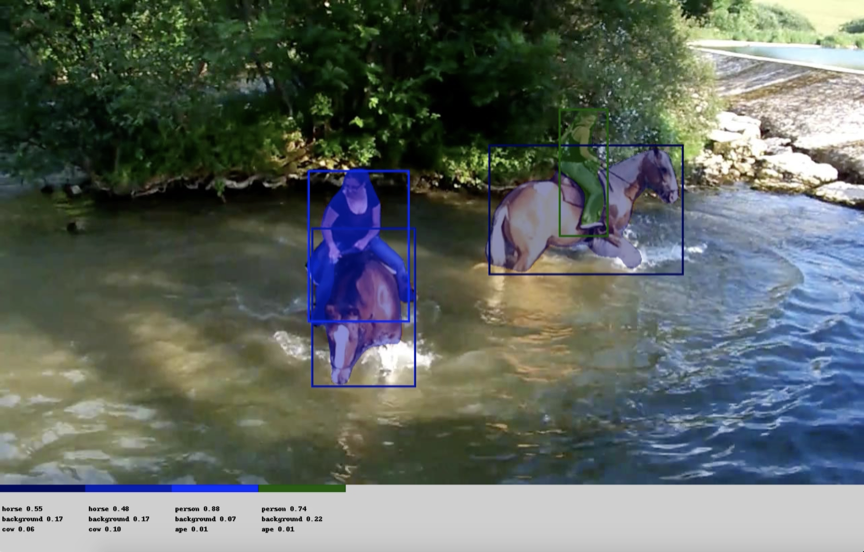}%
  \includegraphics[width=0.25\textwidth,trim={5.5cm 2.0cm 1.0cm 2.0cm}, clip]{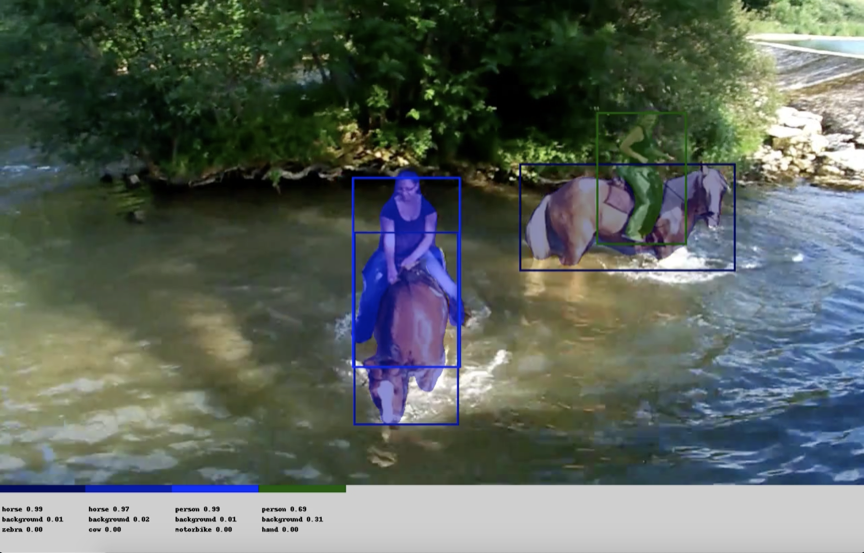}%
  \includegraphics[width=0.25\textwidth,trim={5.5cm 2.0cm 1.0cm 2.0cm}, clip]{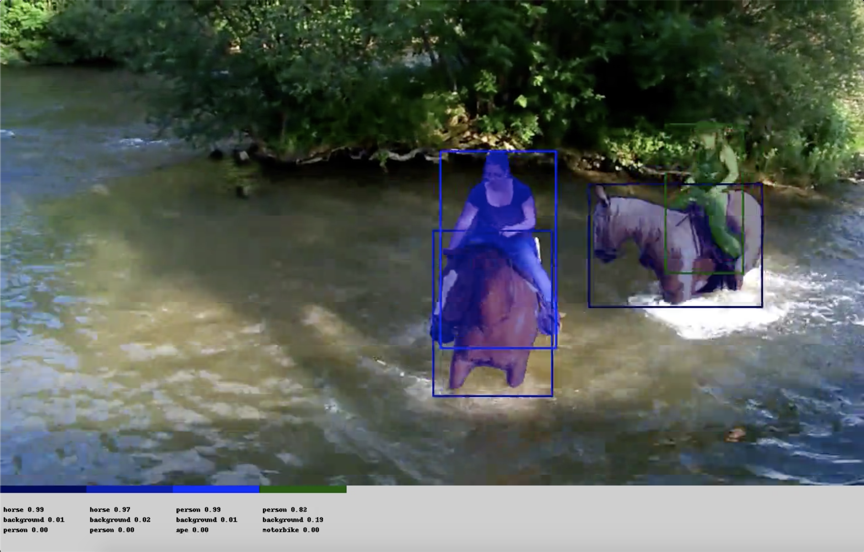}%
  \includegraphics[width=0.25\textwidth,trim={5.5cm 2.0cm 1.0cm 2.0cm}, clip]{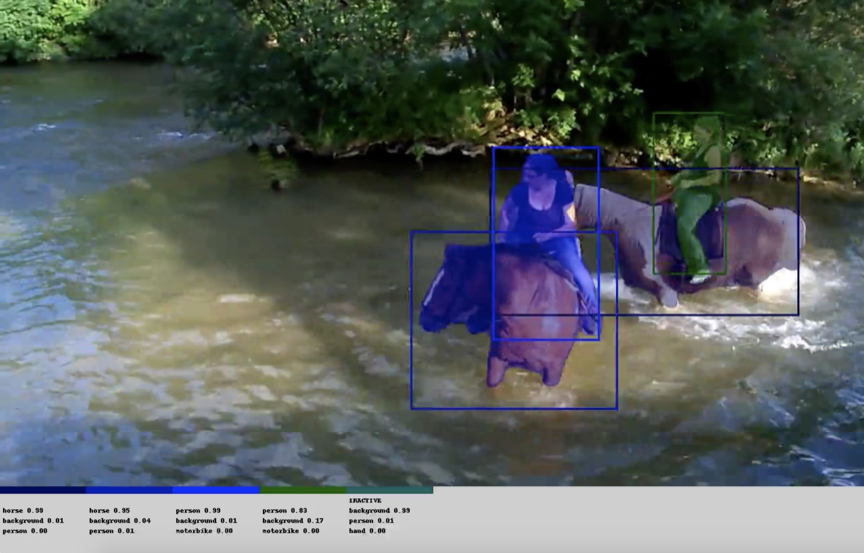}
  \caption{Two example sequences. Our model successfully performs track initialization and track-detection assignment.}
  \label{fig:exp_success}
  \vspace{-3mm}
\end{figure*}


\parsection{Scoring} We predict class membership and confidence based on the track embeddings. Here we try two alternative strategies. First, we try a simple strategy utilized in other works \cite{yang2019video}. The confidence is found as an average of the confidence of all detections assigned to a given track. The class is found via a majority vote. Next, we try to average the scores obtained from YOLACT, which includes both the confidence and class in a single vector. Both these configurations lead to a large drop in performance.

\parsection{LSTM-like gating} We experiment with the LSTM-like gating mechanism. We first try to remove it, directly feeding the track embeddings output from the GNN as input for the subsequent frame. We found that this configuration leads to unstable training and in all attempts diverge, even with gradient clipping and decreased learning rate. We therefore also experiment with a simpler mechanism, adding only a single gate and a $\text{tanh}$ activation. While this setting leads to stabler training, it provides deteriorated performance.

\parsection{Stronger backbone} We opt for real-time performance and therefore selected a lightweight backbone. Here we experiment with replacing our ResNet50 backbone for a ResNet101. This leads to a significant performance boost at additional computational cost.

\subsection{State-of-the-Art Comparison}
We compare our approach to the state-of-the-art, including the baselines proposed in \cite{yang2019video}, and show the results in Table~\ref{tab:quantsota}. Our approach, running at 30 fps, outperforms all near real-time methods. DeepSORT \cite{wojke2017simple}, which relies on Kalman-filtering the bounding boxes and a learnt appearance descriptor used for re-identification, obtains an mAP score of 26.1. MaskTrack R-CNN~\cite{yang2019video} gets a score of 30.3. SipMask~\cite{cao2020sipmask} improves MaskTrack R-CNN by changing its detector and reach a score of 33.7. Using a ResNet50 backbone, we run at similar speed and outperform all three methods with an absolute gain of 9.2, 5.0, and 1.6 respectively.

While \cite{luiten2019video,bertasius2020classifying} obtain higher mAP, those methods are more than a magnitude slower, and thus infeasible for real-time applications or for processing large amounts of data. STEm-Seg~\cite{athar2020stem} reports results using both a ResNet50 and a ResNet101 backbone. We show a gain of 4.7 mAP with ResNet50, and a gain of 3.1 mAP with ResNet101.

\subsection{Qualitative Experiments}
\begin{figure}[b]
  \vspace{-4mm}
  \centering
  \includegraphics[width=0.475\textwidth,trim={0.0cm 0.0cm 0.0cm 0.0cm}, clip]{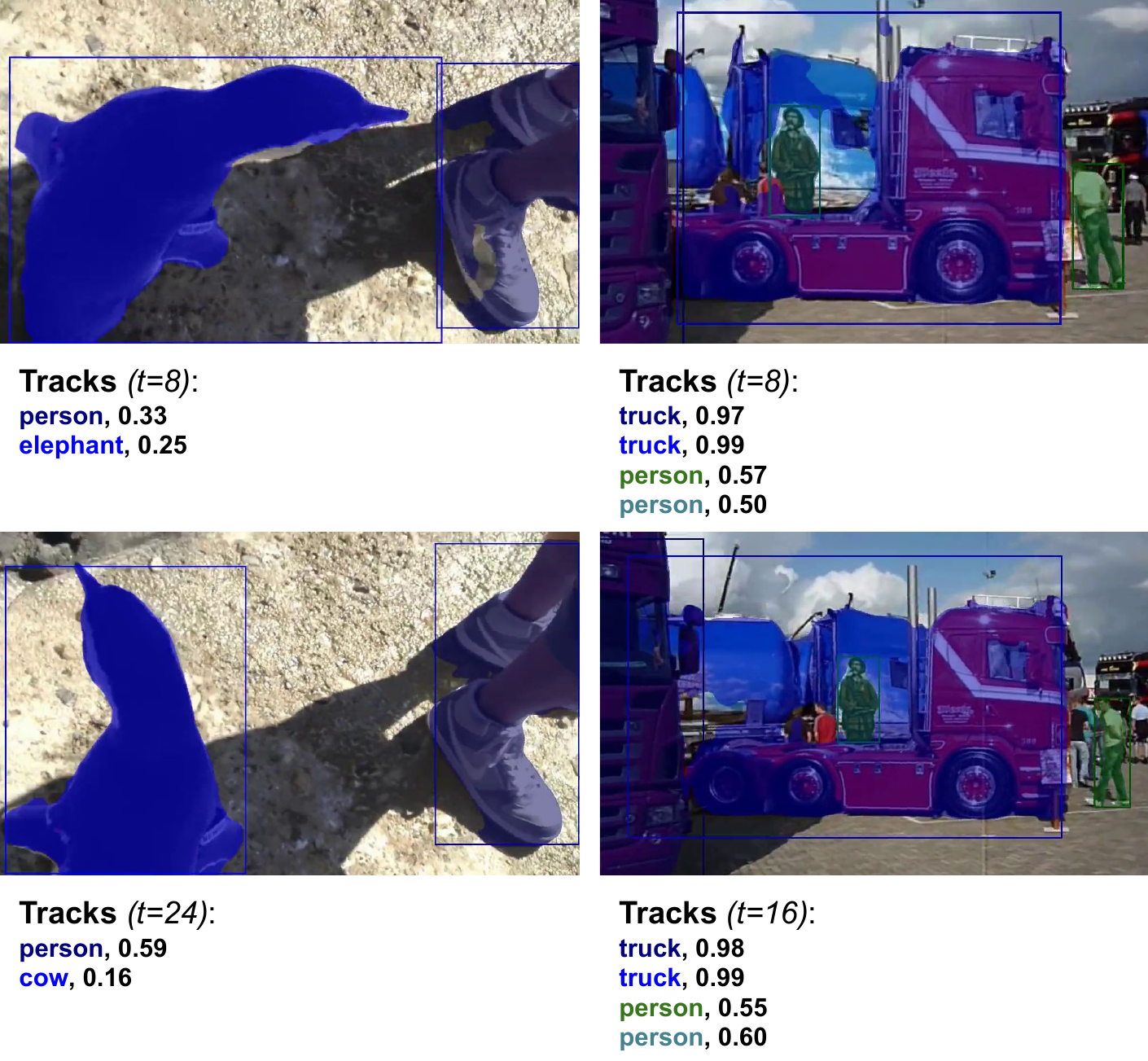}
  \caption{Two failure case examples. A penguin is detected and tracked (left). Penguin is not an annotated class in YouTubeVIS, but our approach deems it similar to an elephant or a cow. To the right, a depiction of a man is detected and tracked, whereas the model misses the truck on which it is painted.}
  \label{fig:exp_fc}
\end{figure}

We provide qualitative results in Figures~\ref{fig:exp_3in1},~\ref{fig:exp_success} and~\ref{fig:exp_fc}. In Figure~\ref{fig:exp_3in1}, we show that our approach is able to handle detection failures: suppressing detector noise, pausing tracks during missed detections, and dealing with false positives. In the latter scenario, a track may initialized for the false positive but its confidence is later reduced and the model ceases to provide it with detections.

In Figure~\ref{fig:exp_success} we show some examples of successful track initialization and track-detection association under the presence of multiple similar objects. Also see Figure~\ref{fig:illustration}. We show example failure cases in Figure~\ref{fig:exp_fc}. The model detects and tracks a penguin even though \emph{penguin} is not an annotated class. Further, the model detects and tracks depictions or pictures of objects.

\section{Conclusion}
We have introduced a novel learning formulation together with an intuitive and flexible model for video instance segmentation. The model proceeds frame by frame, uses as input the detections produced by an instance segmentation method, and incrementally forms tracks. It assigns detections to existing tracks, initializes new tracks, and updates class and confidence in existing tracks. We demonstrate via qualitative and quantitative experiments that the model learns to create accurate tracks, and provide an analysis of its various aspects via ablation experiments.

\textbf{Acknowledgement}: This work was partially supported by the ETH Z\"urich Fund (OK); a Huawei Technologies Oy (Finland) project; ELLIIT; and the Wallenberg AI, Autonomous Systems and Software Program (WASP) funded by the Knut and Alice Wallenberg Foundation.

{\small
\bibliographystyle{ieee_fullname}
\bibliography{references}

\begin{thebibliography}{10}\itemsep=-1pt

\bibitem{athar2020stem}
Ali Athar, Sabarinath Mahadevan, Aljo{\v{s}}a O{\v{s}}ep, Laura Leal-Taix{\'e},
  and Bastian Leibe.
\newblock Stem-seg: Spatio-temporal embeddings for instance segmentation in
  videos.
\newblock In {\em ECCV}, 2020.

\bibitem{DBLP:journals/corr/abs-1806-01261}
Peter~W. Battaglia, Jessica~B. Hamrick, Victor Bapst, Alvaro
  Sanchez{-}Gonzalez, Vin{\'{\i}}cius~Flores Zambaldi, Mateusz Malinowski,
  Andrea Tacchetti, David Raposo, Adam Santoro, Ryan Faulkner, {\c{C}}aglar
  G{\"{u}}l{\c{c}}ehre, H.~Francis Song, Andrew~J. Ballard, Justin Gilmer,
  George~E. Dahl, Ashish Vaswani, Kelsey~R. Allen, Charles Nash, Victoria
  Langston, Chris Dyer, Nicolas Heess, Daan Wierstra, Pushmeet Kohli, Matthew
  Botvinick, Oriol Vinyals, Yujia Li, and Razvan Pascanu.
\newblock Relational inductive biases, deep learning, and graph networks.
\newblock {\em CoRR}, abs/1806.01261, 2018.

\bibitem{OpenImagesSegmentation}
Rodrigo Benenson, Stefan Popov, and Vittorio Ferrari.
\newblock Large-scale interactive object segmentation with human annotators.
\newblock In {\em CVPR}, 2019.

\bibitem{berg2019semi}
Amanda Berg, Joakim Johnander, Flavie Durand~de Gevigney, Jorgen Ahlberg, and
  Michael Felsberg.
\newblock Semi-automatic annotation of objects in visual-thermal video.
\newblock In {\em Proceedings of the IEEE International Conference on Computer
  Vision Workshops}, pages 0--0, 2019.

\bibitem{berman2018lovasz}
Maxim Berman, Amal Rannen~Triki, and Matthew~B Blaschko.
\newblock The lov{\'a}sz-softmax loss: A tractable surrogate for the
  optimization of the intersection-over-union measure in neural networks.
\newblock In {\em Proceedings of the IEEE Conference on Computer Vision and
  Pattern Recognition}, pages 4413--4421, 2018.

\bibitem{bertasius2020classifying}
Gedas Bertasius and Lorenzo Torresani.
\newblock Classifying, segmenting, and tracking object instances in video with
  mask propagation.
\newblock In {\em Proceedings of the IEEE/CVF Conference on Computer Vision and
  Pattern Recognition}, pages 9739--9748, 2020.
\newblock \url{https://arxiv.org/pdf/1912.04573.pdf}.

\bibitem{bertasius2018object}
Gedas Bertasius, Lorenzo Torresani, and Jianbo Shi.
\newblock Object detection in video with spatiotemporal sampling networks.
\newblock In {\em Proceedings of the European Conference on Computer Vision
  (ECCV)}, pages 331--346, 2018.

\bibitem{bolya2019yolact}
Daniel Bolya, Chong Zhou, Fanyi Xiao, and Yong~Jae Lee.
\newblock Yolact: Real-time instance segmentation.
\newblock In {\em Proceedings of the IEEE international conference on computer
  vision}, pages 9157--9166, 2019.

\bibitem{braso2020learning}
Guillem Bras{\'o} and Laura Leal-Taix{\'e}.
\newblock Learning a neural solver for multiple object tracking.
\newblock In {\em Proceedings of the IEEE/CVF Conference on Computer Vision and
  Pattern Recognition}, pages 6247--6257, 2020.
\newblock \url{https://arxiv.org/pdf/1912.07515.pdf}.

\bibitem{burghardt2006analysing}
Tilo Burghardt and J {\'C}ali{\'c}.
\newblock Analysing animal behaviour in wildlife videos using face detection
  and tracking.
\newblock {\em IEE Proceedings-Vision, Image and Signal Processing},
  153(3):305--312, 2006.

\bibitem{cao2020sipmask}
Jiale Cao, Rao~Muhammad Anwer, Hisham Cholakkal, Fahad~Shahbaz Khan, Yanwei
  Pang, and Ling Shao.
\newblock Sipmask: Spatial information preservation for fast instance
  segmentation.
\newblock {\em Proc. European Conference on Computer Vision}, 2020.

\bibitem{chen2019hybrid}
Kai Chen, Jiangmiao Pang, Jiaqi Wang, Yu Xiong, Xiaoxiao Li, Shuyang Sun,
  Wansen Feng, Ziwei Liu, Jianping Shi, Wanli Ouyang, et~al.
\newblock Hybrid task cascade for instance segmentation.
\newblock In {\em Proceedings of the IEEE conference on computer vision and
  pattern recognition}, pages 4974--4983, 2019.

\bibitem{cho2014properties}
Kyunghyun Cho, Bart van Merri{\"e}nboer, Dzmitry Bahdanau, and Yoshua Bengio.
\newblock On the properties of neural machine translation: Encoder--decoder
  approaches.
\newblock {\em Syntax, Semantics and Structure in Statistical Translation},
  page 103, 2014.

\bibitem{cordts2016cityscapes}
Marius Cordts, Mohamed Omran, Sebastian Ramos, Timo Rehfeld, Markus Enzweiler,
  Rodrigo Benenson, Uwe Franke, Stefan Roth, and Bernt Schiele.
\newblock The cityscapes dataset for semantic urban scene understanding.
\newblock In {\em Proceedings of the IEEE conference on computer vision and
  pattern recognition}, pages 3213--3223, 2016.

\bibitem{everingham2015pascal}
Mark Everingham, SM~Ali Eslami, Luc Van~Gool, Christopher~KI Williams, John
  Winn, and Andrew Zisserman.
\newblock The pascal visual object classes challenge: A retrospective.
\newblock {\em International journal of computer vision}, 111(1):98--136, 2015.

\bibitem{greff2016lstm}
Klaus Greff, Rupesh~K Srivastava, Jan Koutn{\'\i}k, Bas~R Steunebrink, and
  J{\"u}rgen Schmidhuber.
\newblock Lstm: A search space odyssey.
\newblock {\em IEEE transactions on neural networks and learning systems},
  28(10):2222--2232, 2016.

\bibitem{han2016seq}
Wei Han, Pooya Khorrami, Tom~Le Paine, Prajit Ramachandran, Mohammad
  Babaeizadeh, Honghui Shi, Jianan Li, Shuicheng Yan, and Thomas~S Huang.
\newblock Seq-nms for video object detection.
\newblock {\em arXiv preprint arXiv:1602.08465}, 2016.

\bibitem{he2016deep}
Kaiming He, Xiangyu Zhang, Shaoqing Ren, and Jian Sun.
\newblock Deep residual learning for image recognition.
\newblock In {\em Proceedings of the IEEE conference on computer vision and
  pattern recognition}, pages 770--778, 2016.

\bibitem{hochreiter1997lstm}
Sepp Hochreiter and J{\"u}rgen Schmidhuber.
\newblock Long short-term memory.
\newblock {\em Neural computation}, 9(8):1735--1780, 1997.

\bibitem{izquierdo2019prevention}
Rub{\'e}n Izquierdo, A Quintanar, Ignacio Parra, D Fern{\'a}ndez-Llorca, and MA
  Sotelo.
\newblock The prevention dataset: a novel benchmark for prediction of vehicles
  intentions.
\newblock In {\em 2019 IEEE Intelligent Transportation Systems Conference
  (ITSC)}, pages 3114--3121. IEEE, 2019.

\bibitem{OpenImages}
Alina Kuznetsova, Hassan Rom, Neil Alldrin, Jasper Uijlings, Ivan Krasin, Jordi
  Pont-Tuset, Shahab Kamali, Stefan Popov, Matteo Malloci, Alexander
  Kolesnikov, Tom Duerig, and Vittorio Ferrari.
\newblock The open images dataset v4: Unified image classification, object
  detection, and visual relationship detection at scale.
\newblock {\em IJCV}, 2020.

\bibitem{luiten2019video}
Jonathon Luiten, Philip Torr, and Bastian Leibe.
\newblock Video instance segmentation 2019: A winning approach for combined
  detection, segmentation, classification and tracking.
\newblock In {\em Proceedings of the IEEE International Conference on Computer
  Vision Workshops}, pages 0--0, 2019.
\newblock
  \url{http://openaccess.thecvf.com/content_ICCVW_2019/papers/YouTube-VOS/Luiten_Video_Instance_Segmentation_2019_A_Winning_Approach_for_Combined_Detection_ICCVW_2019_paper.pdf}.

\bibitem{luiten2020unovost}
Jonathon Luiten, Idil~Esen Zulfikar, and Bastian Leibe.
\newblock Unovost: Unsupervised offline video object segmentation and tracking.
\newblock In {\em 2020 IEEE Winter Conference on Applications of Computer
  Vision (WACV)}, pages 1989--1998. IEEE, 2020.

\bibitem{murphy2007conjugate}
Kevin~P Murphy.
\newblock Conjugate bayesian analysis of the gaussian distribution.
\newblock {\em def}, 1(2$\sigma$2):16, 2007.

\bibitem{pytorch}
Adam Paszke, Sam Gross, Soumith Chintala, Gregory Chanan, Edward Yang, Zachary
  DeVito, Zeming Lin, Alban Desmaison, Luca Antiga, and Adam Lerer.
\newblock Automatic differentiation in pytorch.
\newblock 2017.

\bibitem{sarlin2020superglue}
Paul-Edouard Sarlin, Daniel DeTone, Tomasz Malisiewicz, and Andrew Rabinovich.
\newblock Superglue: Learning feature matching with graph neural networks.
\newblock In {\em Proceedings of the IEEE/CVF Conference on Computer Vision and
  Pattern Recognition}, pages 4938--4947, 2020.

\bibitem{t2016automatic}
Roeland T'Jampens, Francisco Hernandez, Florian Vandecasteele, and Steven
  Verstockt.
\newblock Automatic detection, tracking and counting of birds in marine video
  content.
\newblock In {\em 2016 Sixth International Conference on Image Processing
  Theory, Tools and Applications (IPTA)}, pages 1--6. IEEE, 2016.

\bibitem{voigtlaender2019feelvos}
Paul Voigtlaender, Yuning Chai, Florian Schroff, Hartwig Adam, Bastian Leibe,
  and Liang-Chieh Chen.
\newblock Feelvos: Fast end-to-end embedding learning for video object
  segmentation.
\newblock In {\em Proceedings of the IEEE Conference on Computer Vision and
  Pattern Recognition}, pages 9481--9490, 2019.

\bibitem{weng2020gnn3dmot}
Xinshuo Weng, Yongxin Wang, Yunze Man, and Kris~M. Kitani.
\newblock {GNN3DMOT:} graph neural network for 3d multi-object tracking with
  2d-3d multi-feature learning.
\newblock In {\em 2020 {IEEE/CVF} Conference on Computer Vision and Pattern
  Recognition, {CVPR} 2020, Seattle, WA, USA, June 13-19, 2020}, pages
  6498--6507. {IEEE}, 2020.

\bibitem{wojke2017simple}
Nicolai Wojke, Alex Bewley, and Dietrich Paulus.
\newblock Simple online and realtime tracking with a deep association metric.
\newblock In {\em 2017 IEEE international conference on image processing
  (ICIP)}, pages 3645--3649. IEEE, 2017.

\bibitem{yang2019video}
Linjie Yang, Yuchen Fan, and Ning Xu.
\newblock Video instance segmentation.
\newblock In {\em Proceedings of the IEEE International Conference on Computer
  Vision}, pages 5188--5197, 2019.
\newblock \url{https://arxiv.org/pdf/1905.04804.pdf}.

\bibitem{OSMN}
Linjie Yang, Yanran Wang, Xuehan Xiong, Jianchao Yang, and Aggelos~K.
  Katsaggelos.
\newblock Efficient video object segmentation via network modulation.
\newblock In {\em The IEEE Conference on Computer Vision and Pattern
  Recognition (CVPR)}, June 2018.

\bibitem{yu2020bdd100k}
Fisher Yu, Haofeng Chen, Xin Wang, Wenqi Xian, Yingying Chen, Fangchen Liu,
  Vashisht Madhavan, and Trevor Darrell.
\newblock Bdd100k: A diverse driving dataset for heterogeneous multitask
  learning.
\newblock In {\em Proceedings of the IEEE/CVF Conference on Computer Vision and
  Pattern Recognition}, pages 2636--2645, 2020.

\bibitem{zhang2008automatic}
Xiao-yan Zhang, Xiao-juan Wu, Xin Zhou, Xiao-gang Wang, and Yuan-yuan Zhang.
\newblock Automatic detection and tracking of maneuverable birds in videos.
\newblock In {\em 2008 International Conference on Computational Intelligence
  and Security}, volume~1, pages 185--189. IEEE, 2008.

\end{thebibliography}
}

\newpage
\appendix
\section*{Appendix}
\renewcommand*{\thesection}{\Alph{section}}
In this appendix, we first provide the details of our proposed model. Next, we present a deeper analysis of our approach by providing more details to the ablative study presented in the main paper. We then go through details on our training data, and last we show additional qualitative results.

\section{Model Details}
We supply additional details on the training of our model, the model architecture, and how we deal with the sparsity of the tracks and detections.

\subsection{Model Training Stages}
While it is theoretically possible to train our model end-to-end in a single stage, we instead opt for a greedy approach and where the model is trained in two stages. We first train the instance segmentation method and then we train all other components. There are two reasons for this: (i) the instance segmentation method does not benefit from replacing single images with videos and doing so only makes training slower; and (ii) training the entire model together, including the instance segmentation method, on a batch of video clips is prohibitively expensive in terms of memory consumption. We therefore first train the backbone and the instance segmentation method on single images for 120 epochs. In this stage, we let the batch normalization layers in the backbone keep running averages of the batch statistics. We then train all other components for 150 epochs: the appearance network, the graph neural network (GNN), the mask weighting module, the recurrent gating mechanism, the logistic model for track-detection matching, the track logistic model for track initialization, the track scoring multinomial logistic model, and the appearance update predictors. During this second training stage, the backbone and the instance segmentation are frozen, including the batch statistics of the former.

\subsection{Architecture Details}
We provide the details on the architecture of our approach. We use a ResNet50 or a ResNet101 backbone~\cite{he2016deep} and the instance segmentation method YOLACT~\cite{bolya2019yolact}. We use dropout regularization, randomly dropping channels with probability $0.1$ on all feature maps extracted by the backbone before feeding them into the instance segmentation method and our appearance network. For the post-processing inside YOLACT, we keep detections with a confidence of at least $0.03$ for any class, and set the non-maximum suppression intersection over union threshold to $0.7$. We found that these values led to good performance of our final method on the 200 videos we held out for validation. We supply the exact architecture of our appearance network, graph neural network (GNN), and mask reweighting module in Table~\ref{tab:architecture}.

\subsection{Dealing with Sparsity}
The tracks and detections vary in number between frames in a single sequence, and between training examples in a batch. We use tensors of fixed size together with a mask, marking which elements are active and which are not. During for instance aggregation in the nodes, the representation is masked with zeros for inactive edges. We set the maximum size for these tensors to $24$ tracks and $16$ detections. 

\begin{table}
  \centering%
  \resizebox{\columnwidth}{!}{%
    \begin{tabular}{|l|l|}\hline
      \multicolumn{2}{|c|}{\textbf{Appearance Network}}\\
      processes conv4\_x & \texttt{Conv2d(1024, 256, 1)}\\
      processes conv5\_x & \texttt{Conv2d(2048, 256, 1)}\\
      & mask-pool and concatenate\\
      & \texttt{Residual(512, 128)}\\\hline
      \multicolumn{2}{|c|}{\textbf{GNN block 1}}\\
      $f^e$ & \texttt{Linear(175, 128) + ReLU + Residual(128, 32)}\\
      $g^\tau$ & \texttt{Linear(128, 32) + ReLU + Linear(32, 128) + Sigmoid}\\
      $f^\tau$ & \texttt{Linear(256, 128) + ReLU}\\
      $g^\tau$ & \texttt{Linear(128, 32) + ReLU + Linear(32, 128) + Sigmoid}\\
      $f^\tau$ & \texttt{Linear(256, 128) + ReLU}\\
      $g^\delta$ & \texttt{Linear(128, 32) + ReLU + Linear(32, 128) + Sigmoid}\\
      $f^\delta$ & \texttt{Linear(173, 128) + ReLU}\\\hline
      \multicolumn{2}{|c|}{\textbf{GNN Residual block}}\\
      processes edges & \texttt{Residual(128, 32)}\\
      processes tracks & \texttt{Residual(128, 32)}\\
      processes detections & \texttt{Residual(128, 32)}\\\hline
      \multicolumn{2}{|c|}{\textbf{GNN block 2}}\\
      $f^e$ & \texttt{Linear(384, 128) + ReLU}\\
      $g^\tau$ & \texttt{Linear(128, 32) + ReLU + Linear(32, 128) + Sigmoid}\\
      $f^\tau$ & \texttt{Linear(256, 128) + ReLU}\\
      $g^\tau$ & \texttt{Linear(128, 32) + ReLU + Linear(32, 128) + Sigmoid}\\
      $f^\tau$ & \texttt{Linear(256, 128) + ReLU}\\
      $g^\delta$ & \texttt{Linear(128, 32) + ReLU + Linear(32, 128) + Sigmoid}\\
      $f^\delta$ & \texttt{Linear(256, 128) + ReLU}\\\hline
      \multicolumn{2}{|c|}{\textbf{GNN Residual block}}\\
      processes edges & \texttt{Residual(128, 32)}\\
      processes tracks & \texttt{Residual(128, 32)}\\
      processes detections & \texttt{Residual(128, 32)}\\\hline
      \multicolumn{2}{|c|}{\textbf{Mask weighting module}}\\
      processes tracks & \texttt{Linear(128, 16) + ReLU}\\
      & broadcast to mask size\\
      & concatenate with detection mask and box (as mask)\\
      & \texttt{Conv2d(18, 16, 3) + ReLU + Conv2d(16, 1, 3)}\\\hline
    \end{tabular}
  }
  \vspace{1mm}
  \caption{Pseudo-code for the construction of the different neural networks of our method. conv4\_x and conv5\_x denote the last stride 16 and last stride 32 feature maps of the backbone. \texttt{Conv2d($D^\text{in}$, $D^\text{out}$, $K\times K$)} denotes a 2-dimensional convolutional layer with $D^\text{in}$ channels in, $D^\text{out}$ channels out, and a kernel size of $K$. \texttt{Linear($D^\text{in}$, $D^\text{out}$)} denotes a linear layer. \texttt{Residual($D^\text{in}$, $D^\text{bottleneck}$)} denotes the residual network bottleneck~\cite{he2016deep}, but with linear layers instead of convolutions. The bottleneck comprises three linear layers, the first projecting the input down to $D^\text{bottleneck}$ channels and the last transforming the input back to $D^\text{in}$ channels. \texttt{Sigmoid} is the logistic function and \texttt{ReLU} the rectified linear unit.}
  \label{tab:architecture}
\end{table}

\section{Detailed GNN Experiments}
We supply four experiments with the graph neural network to provide an even more detailed analysis of our approach. The results are shown in Table~\ref{tab:supplementary:experiments}. Each experiment is detailed as follows.

\begin{table}
  \centering%
  \resizebox{0.6\columnwidth}{!}{%
    \begin{tabular}{l l}\hline
      \textbf{Configuration} & \textbf{mAP}\\\hline
      Proposed approach & 35.3\\\hline
      MLP Node Updates & 34.1\\
      1 GNN Block & 32.4\\
      3 GNN Blocks & 35.2\\
      No Interleaved Residual Blocks & 33.3\\\hline
    \end{tabular}
  }
  \vspace{1mm}
  \caption{Results of five additional experiments on the YouTubeVIS~\cite{yang2019video} validation set.}
  \label{tab:supplementary:experiments}
\end{table}

\paragraph{MLP Node Updates} In our approach we utilize sigmoid gates in the node updates. As mentioned in Section 3.1 in the main paper, the purpose of these gates is to permit a node to dynamically select from which other nodes it wants to gather information. This is in contrast to for instance \cite{braso2020learning} where a 2-layer multilayer perceptron (MLP) is utilized. We experiment with removing the gates and instead adding an extra layer to $f^e$, $f^\tau$, and $f^\delta$, making them into 2-layer MLPs. In doing so, we observe a drop of $1.2$ mAP.


\paragraph{Number of GNN Blocks} As mentioned in Section 3.6 in the main paper, we opted to use two graph blocks. This is the same as the length of the longest path in our bipartite graph, and we are therefore able to feed information from any graph element to any other graph element. We try to instead use a single GNN block and to use three GNN blocks. Note that each GNN block is followed by a residual block, as shown in Table~\ref{tab:architecture}. With a single GNN block, and thus with a model unable to propagate information between any pair of graph elements, we observe a drop of $2.9$ mAP. Using three GNN blocks, compared to using two GNN blocks, leads to a minor drop of $0.1$ mAP.

\paragraph{No Interleaved Residual Blocks} As argued in Section 3.1 in the main paper, we interleave the GNN blocks with residual blocks in order to add some flexibility to the GNN. We try to run without them. This leads to a drop of $2.0$ mAP.

\section{Data Details}
We provide additional details on the datasets used for training, how the data is augmented, and how it is sampled.

\subsection{Datasets}
For the training of the instance segmentation method we use a mix of YouTubeVIS~\cite{yang2019video} and OpenImages~\cite{OpenImages,OpenImagesSegmentation}. The YouTubeVIS training set contains 2238 videos, but we keep only those that are of the correct size, $720\times 1280$, and hold out 200 such sequences for validation, giving us a total of 1867 videos for training. We consider each video a sample, and randomly select a single frame. OpenImages contains 237272 images with instance segmentation annotations for most objects of the YouTubeVIS categories. The images vary in resolution, and we resize them to fit $720\times 1280$. Note that OpenImages is only sparsely annotated, containing annotations for only a fraction of all objects in each image. We do not treat the samples from OpenImages differently however.

\subsection{Data Augmentation}
For each training example, we first uniformly sample height and width from $\mathcal{U}(\begin{bmatrix} 342 & 608 \end{bmatrix}, \begin{bmatrix} 720 & 1280 \end{bmatrix})$ and then randomly crop with zero-padding to get an image of size $480\times 720$. We make sure to remove any objects that have disappeared due to the cropping. Last, we randomly flip the image horizontally. During the second training stage, when we train with sequences, we use the same augmentation for all images in a given sequence.

\subsection{Dataset Sampling}
We train the instance segmentation method for 120 epochs, each comprising 7468 samples, i.e. four times the size of YouTubeVIS. We randomly sample without replacement from the two datasets, weighting each training sample such that there initially is a $75\%$ chance to select an example from OpenImages.

\begin{figure*}[!t]
 \centering
 \includegraphics[width=0.2\textwidth,trim={2.0cm 3.5cm 3.5cm 0.5cm}, clip]{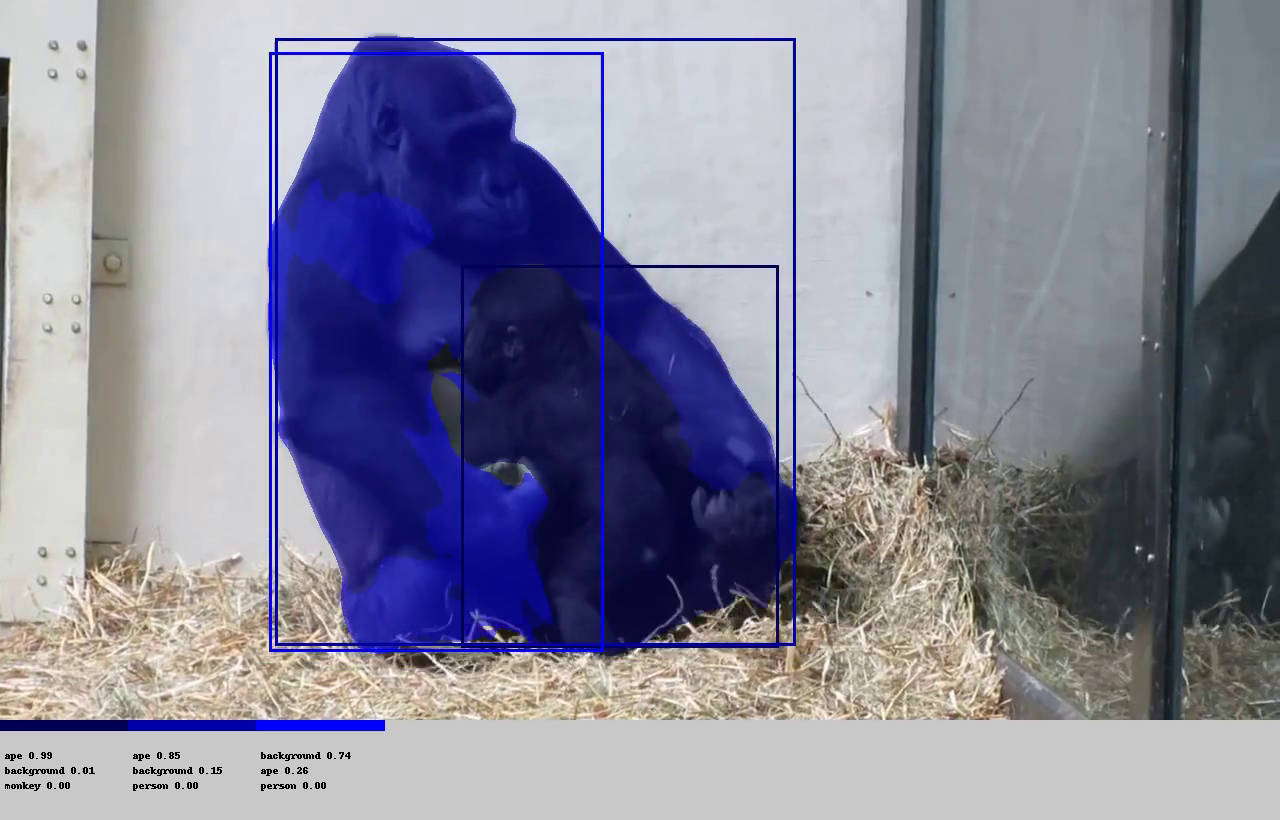}\hspace*{0.2mm}%
 \includegraphics[width=0.2\textwidth,trim={2.0cm 3.5cm 3.5cm 0.5cm}, clip]{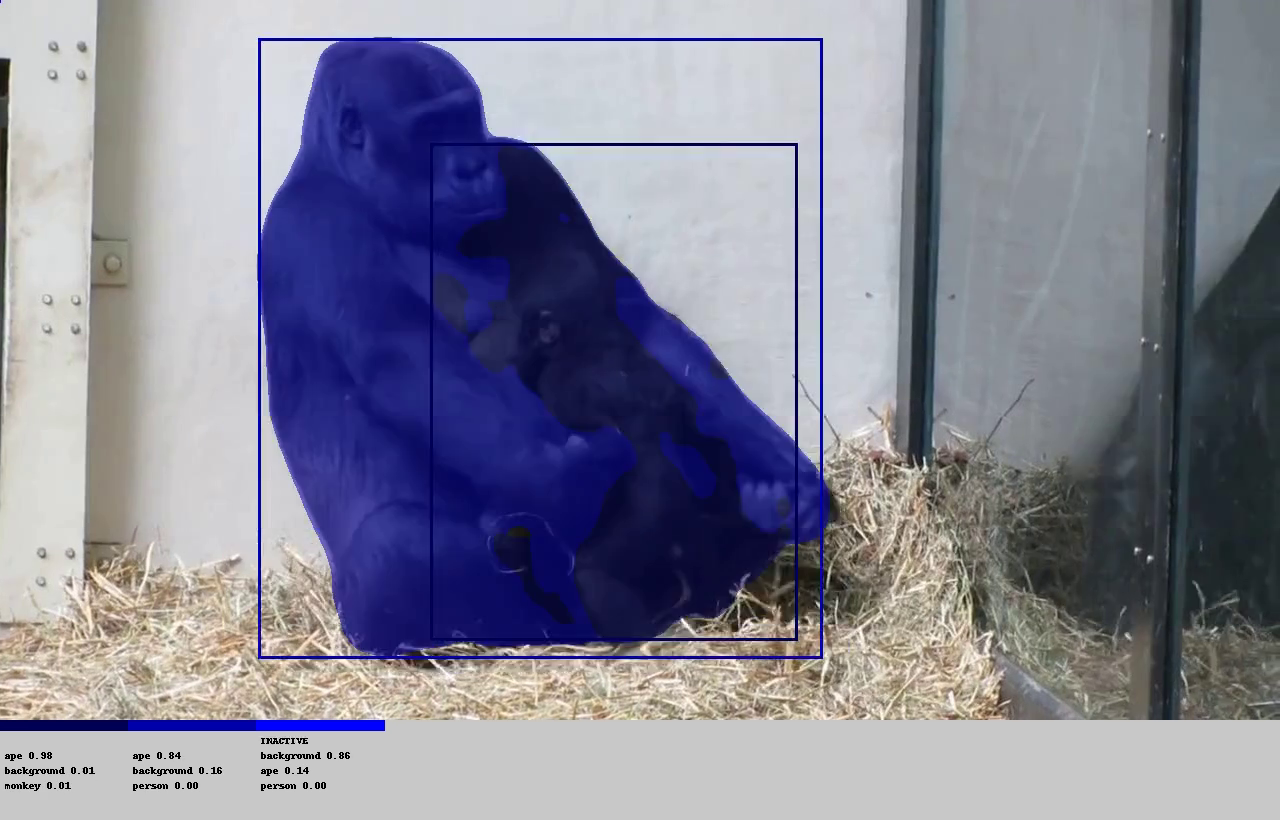}\hspace*{0.2mm}%
 \includegraphics[width=0.2\textwidth,trim={2.0cm 3.5cm 3.5cm 0.5cm}, clip]{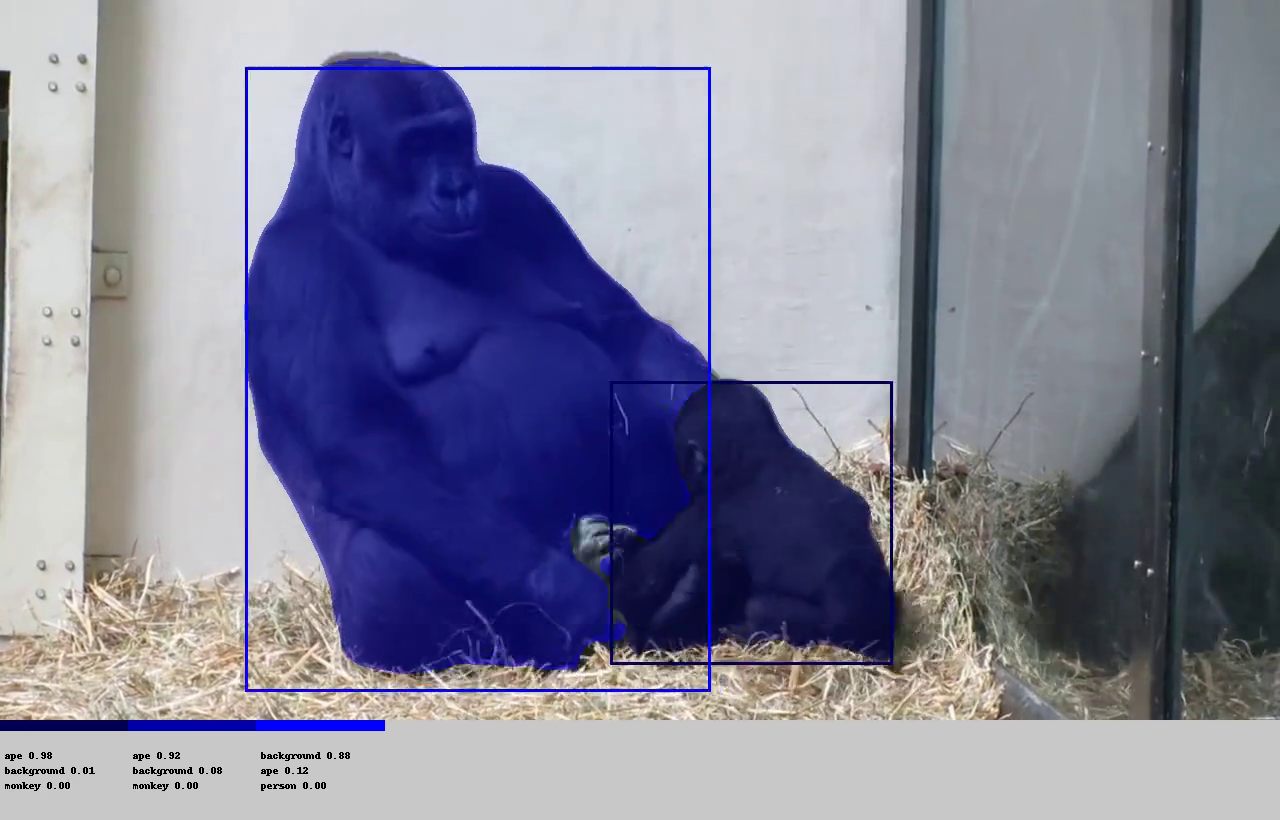}\hspace*{0.2mm}%
 \includegraphics[width=0.2\textwidth,trim={2.0cm 3.5cm 3.5cm 0.5cm}, clip]{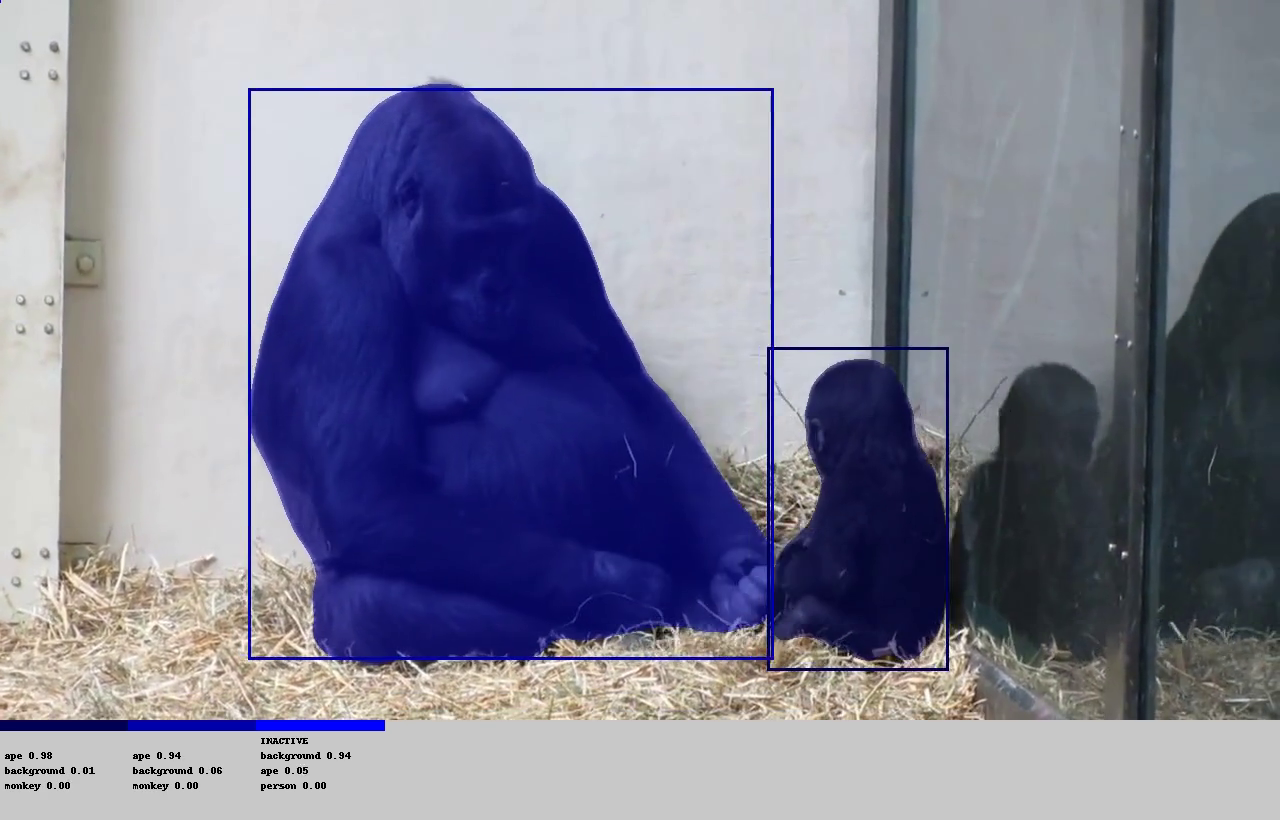}%
 \includegraphics[width=0.19\textwidth,trim={0.0cm 0.5cm 0.0cm 0.0cm}, clip]{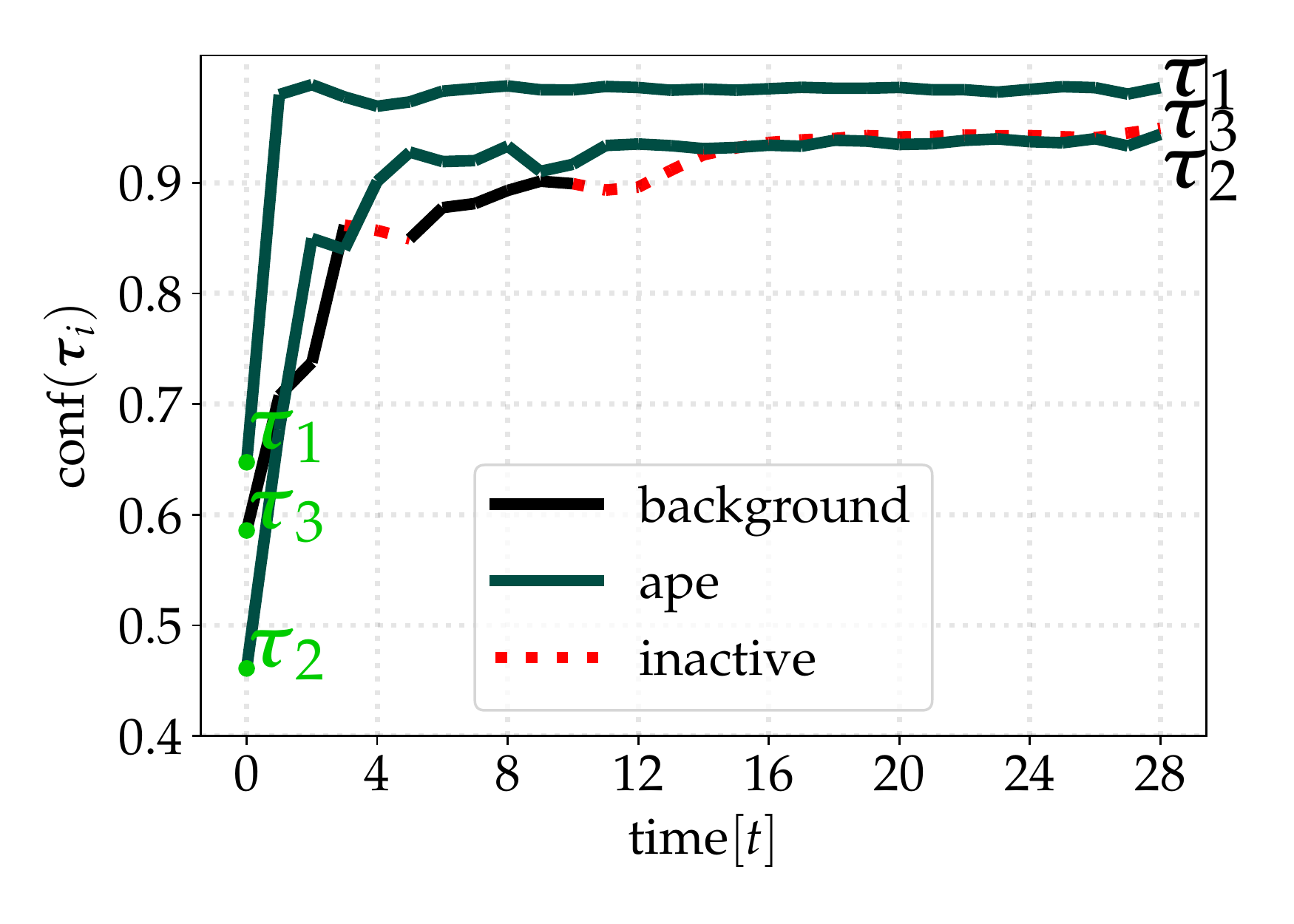} 
 
 \includegraphics[width=0.2\textwidth,trim={2.0cm 3.5cm 3.5cm 0.5cm}, clip]{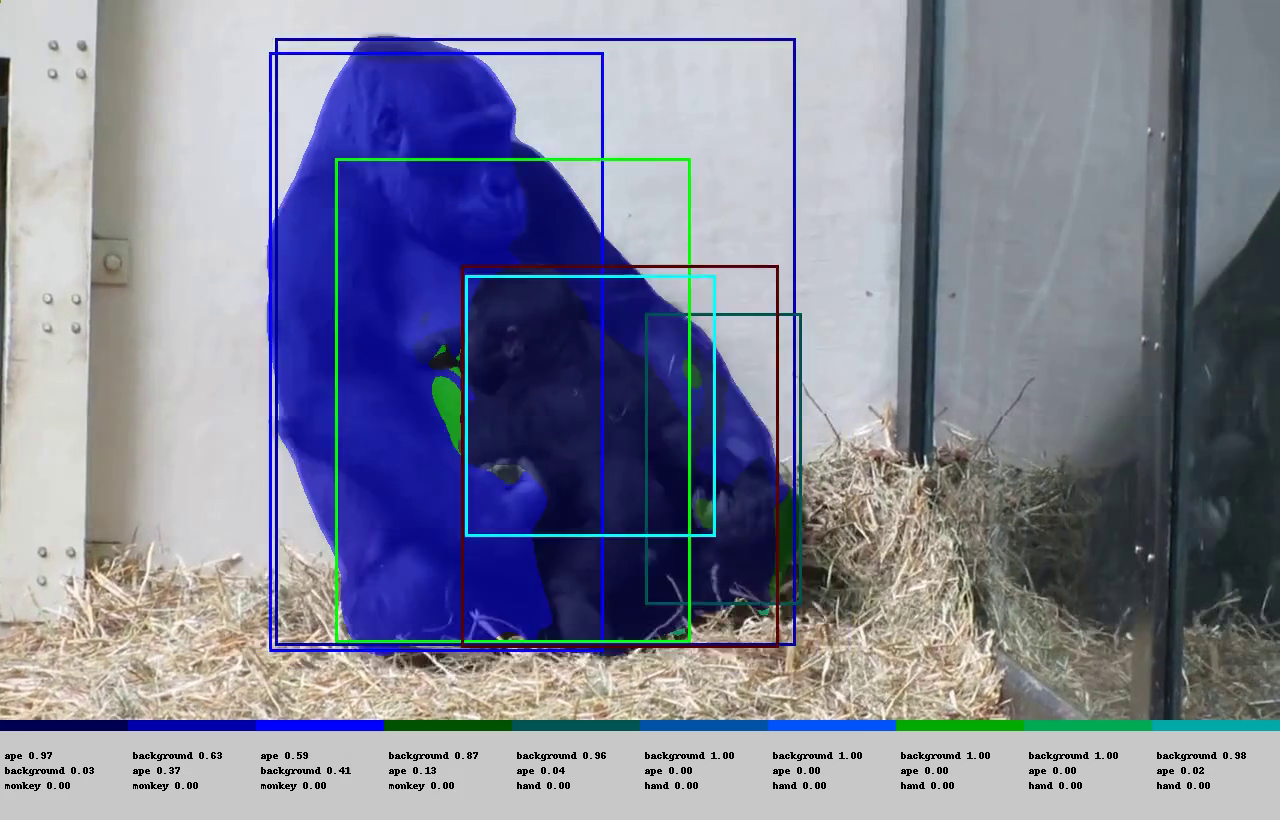}\hspace*{0.2mm}%
 \includegraphics[width=0.2\textwidth,trim={2.0cm 3.5cm 3.5cm 0.5cm}, clip]{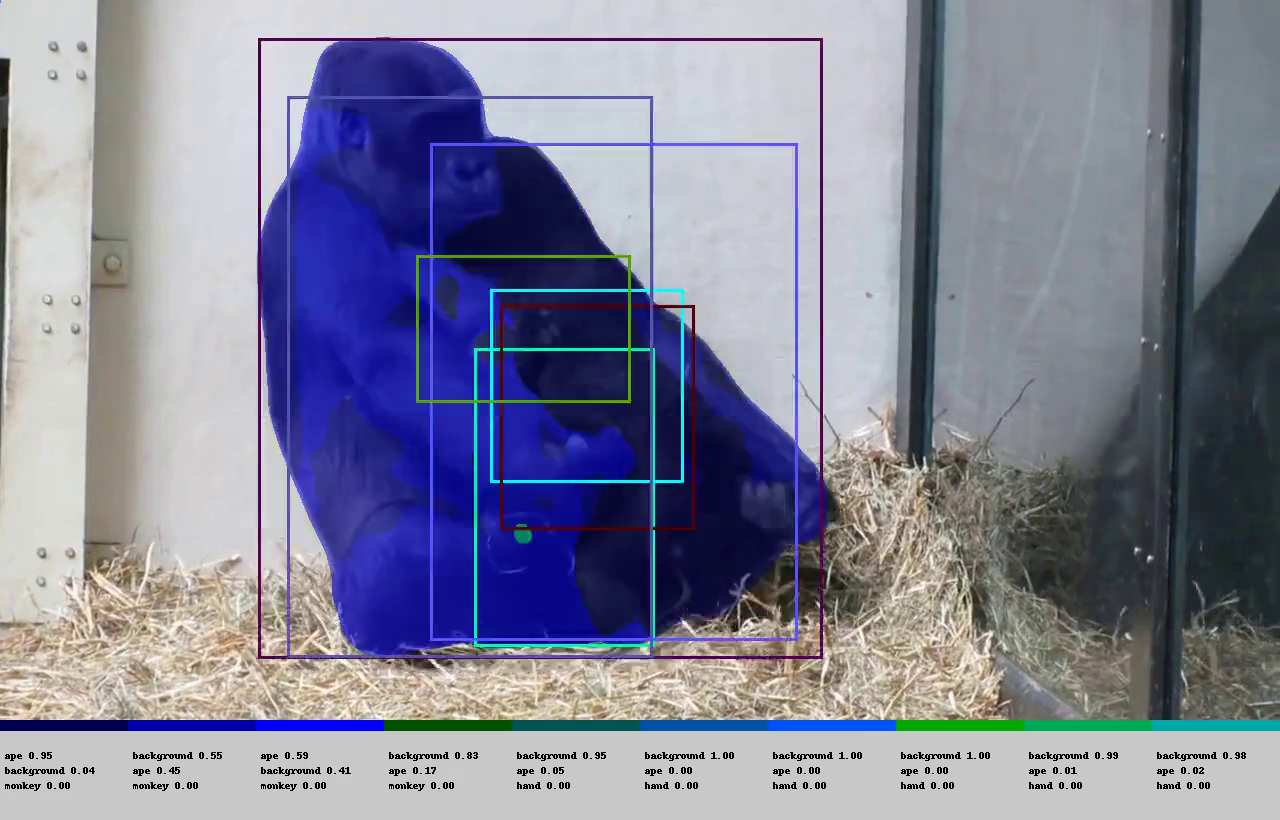}\hspace*{0.2mm}%
 \includegraphics[width=0.2\textwidth,trim={2.0cm 3.5cm 3.5cm 0.5cm}, clip]{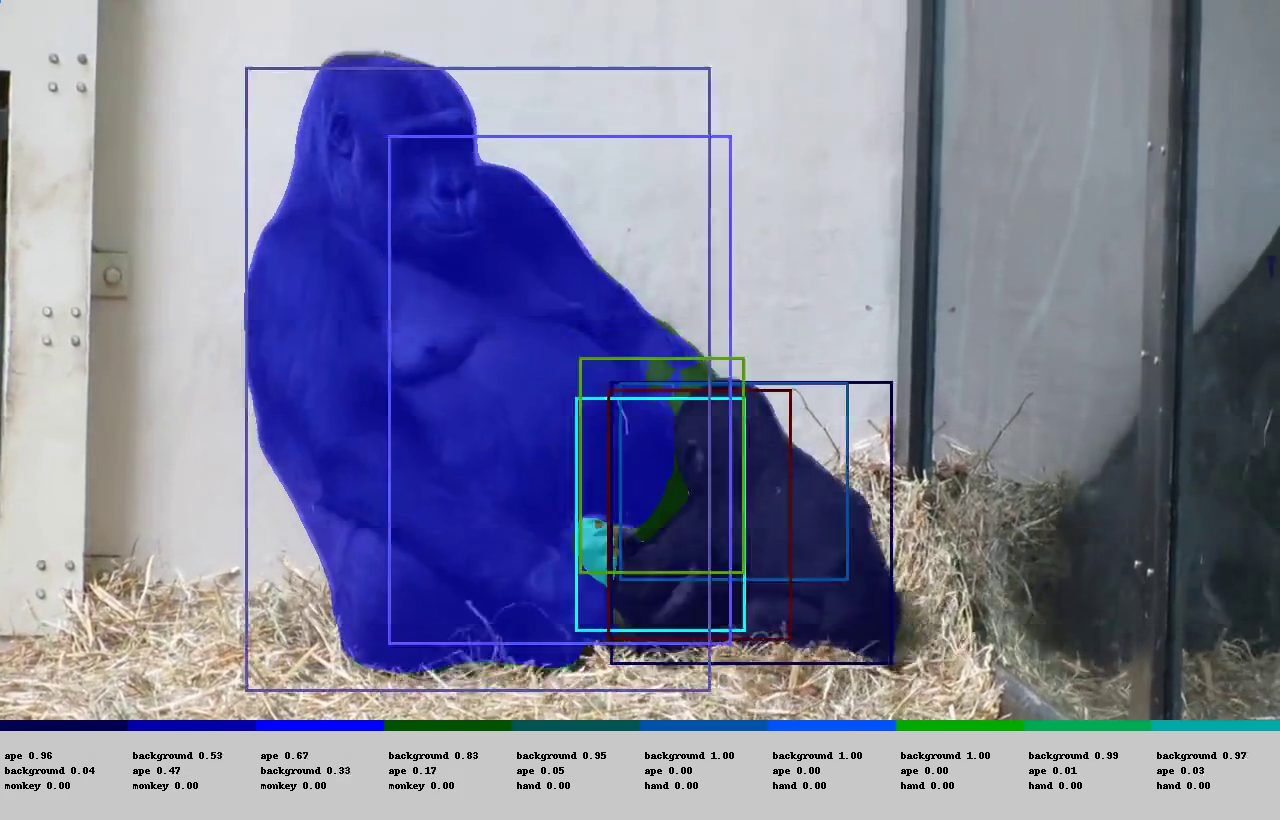}\hspace*{0.2mm}%
 \includegraphics[width=0.2\textwidth,trim={2.0cm 3.5cm 3.5cm 0.5cm}, clip]{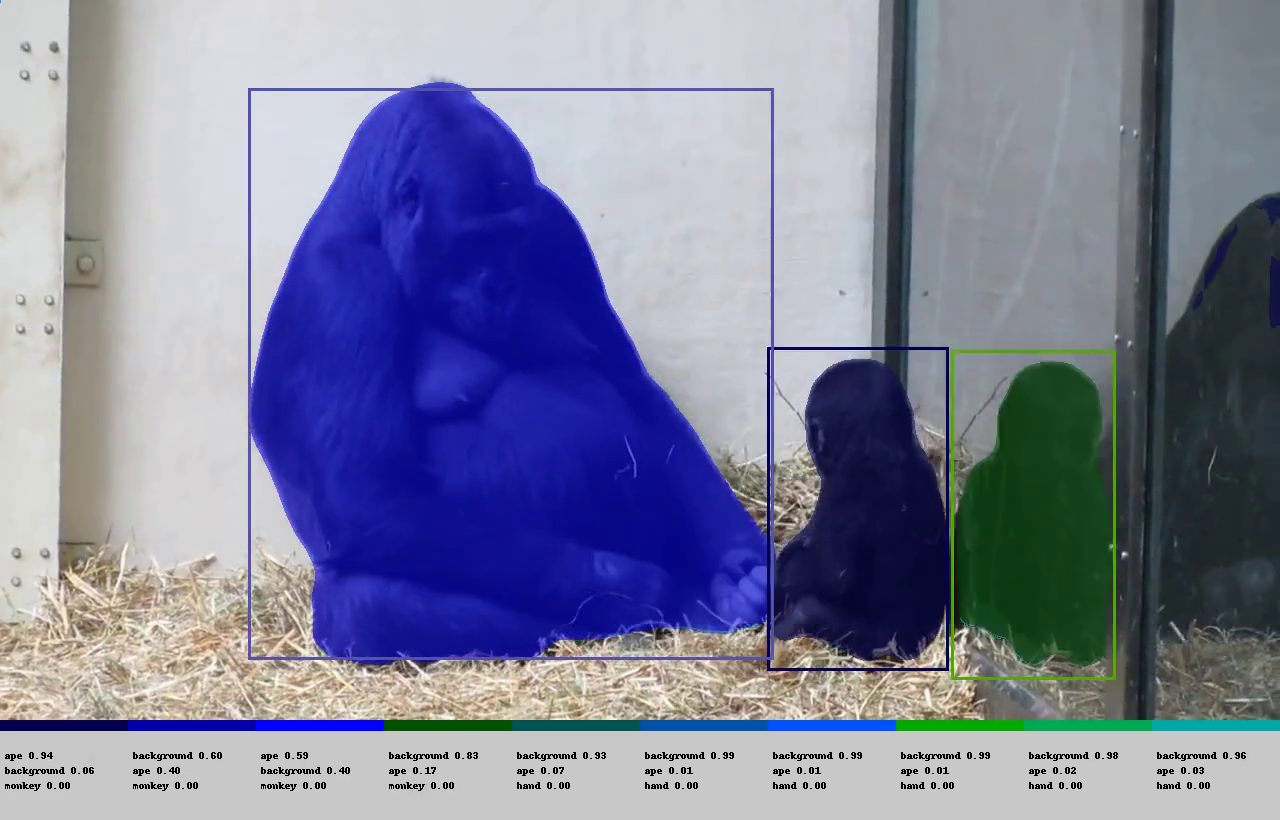}%
 \includegraphics[width=0.19\textwidth,trim={0.0cm 0.5cm 0.0cm 0.0cm}, clip]{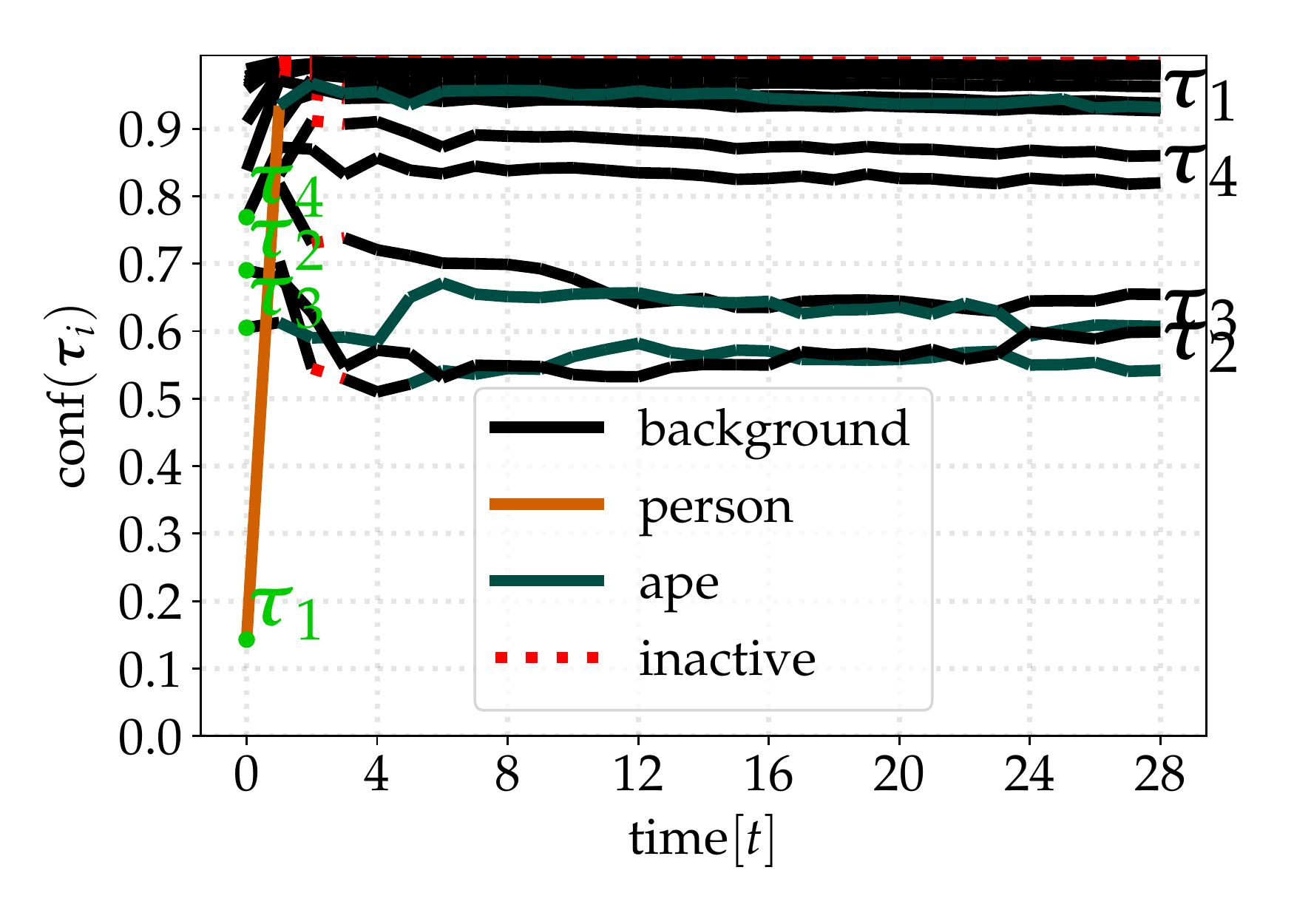} 
 
 \includegraphics[width=0.2\textwidth,trim={2.0cm 3.5cm 3.5cm 0.5cm}, clip]{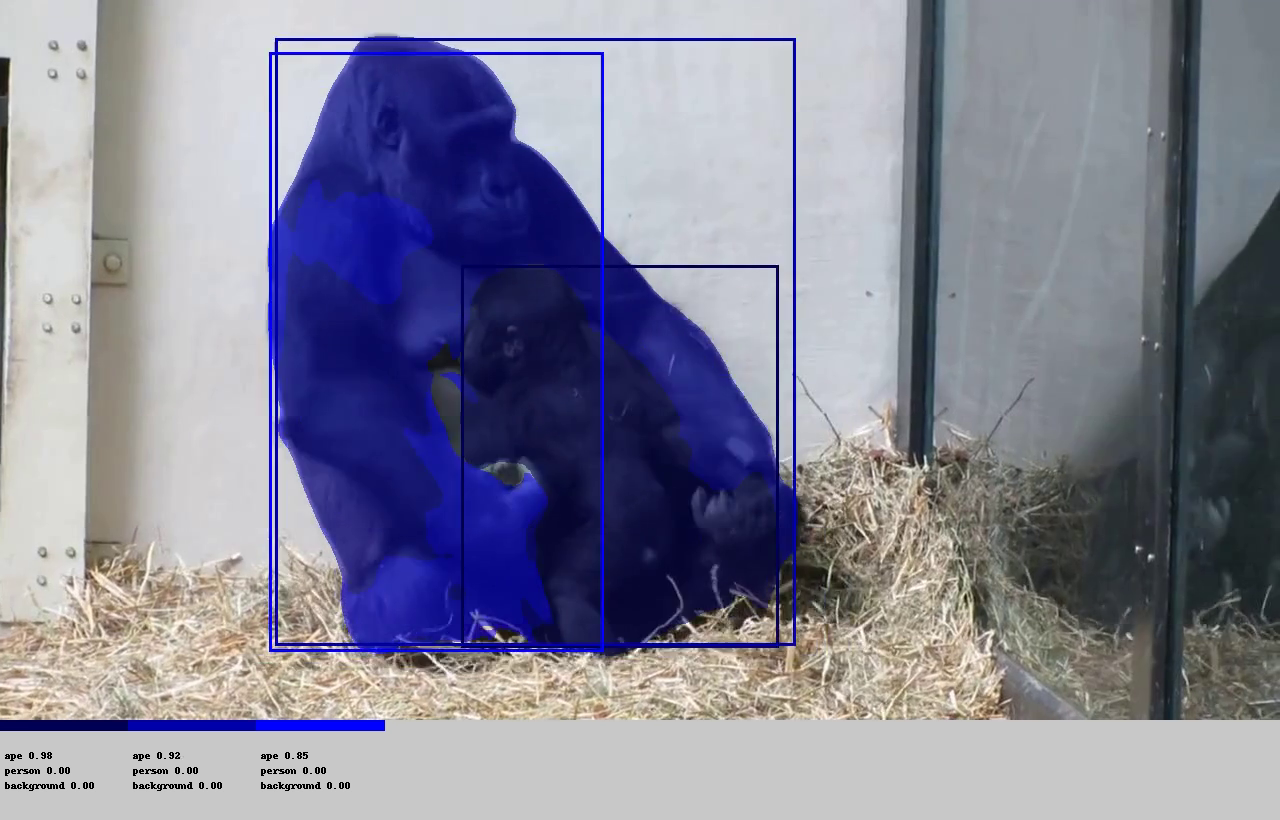}\hspace*{0.2mm}%
 \includegraphics[width=0.2\textwidth,trim={2.0cm 3.5cm 3.5cm 0.5cm}, clip]{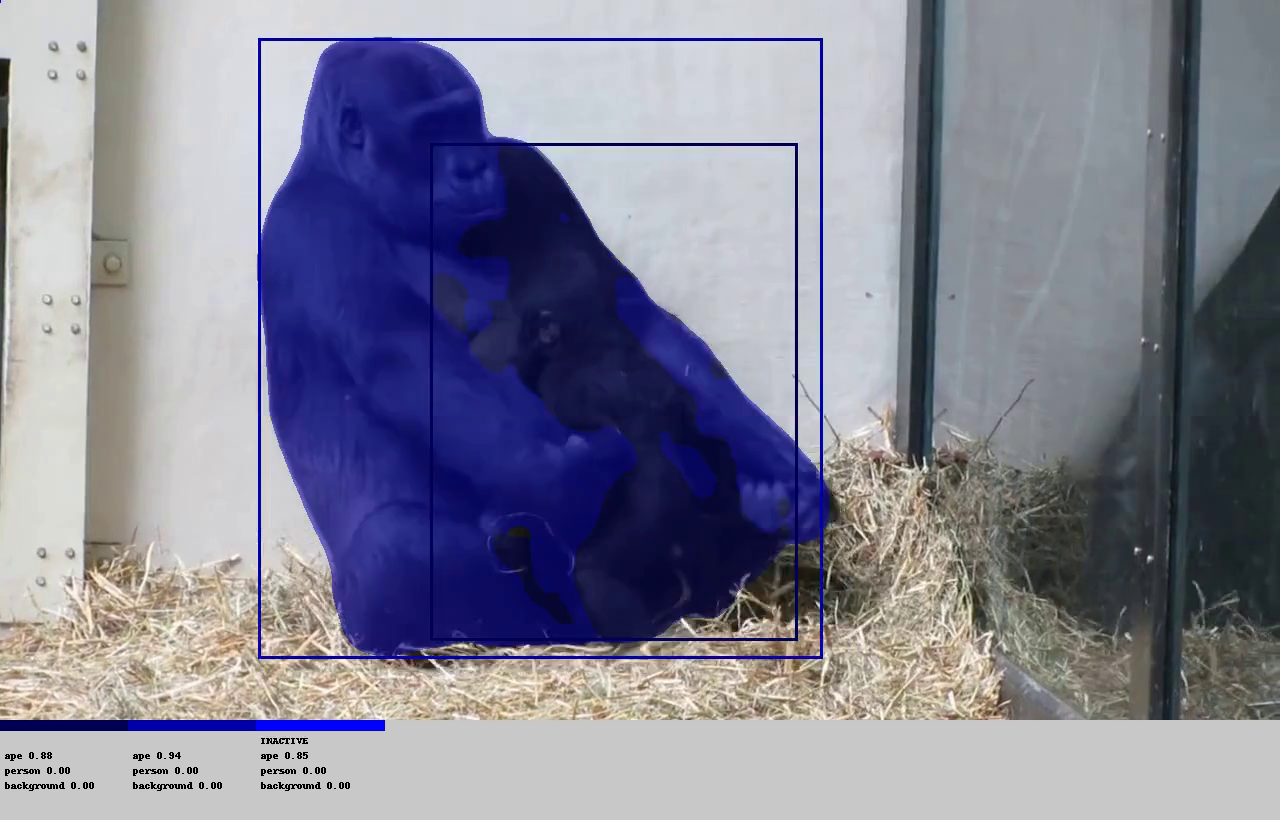}\hspace*{0.2mm}%
 \includegraphics[width=0.2\textwidth,trim={2.0cm 3.5cm 3.5cm 0.5cm}, clip]{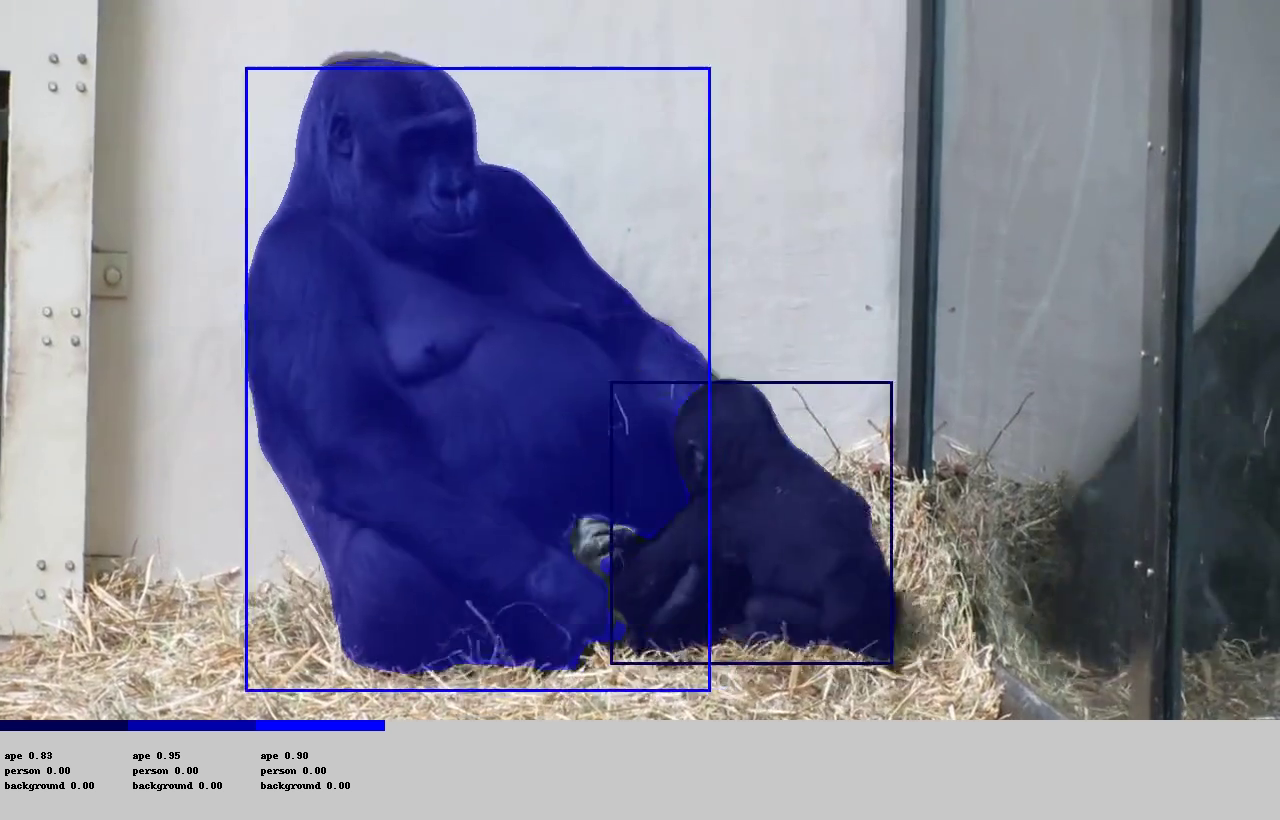}\hspace*{0.2mm}%
 \includegraphics[width=0.2\textwidth,trim={2.0cm 3.5cm 3.5cm 0.5cm}, clip]{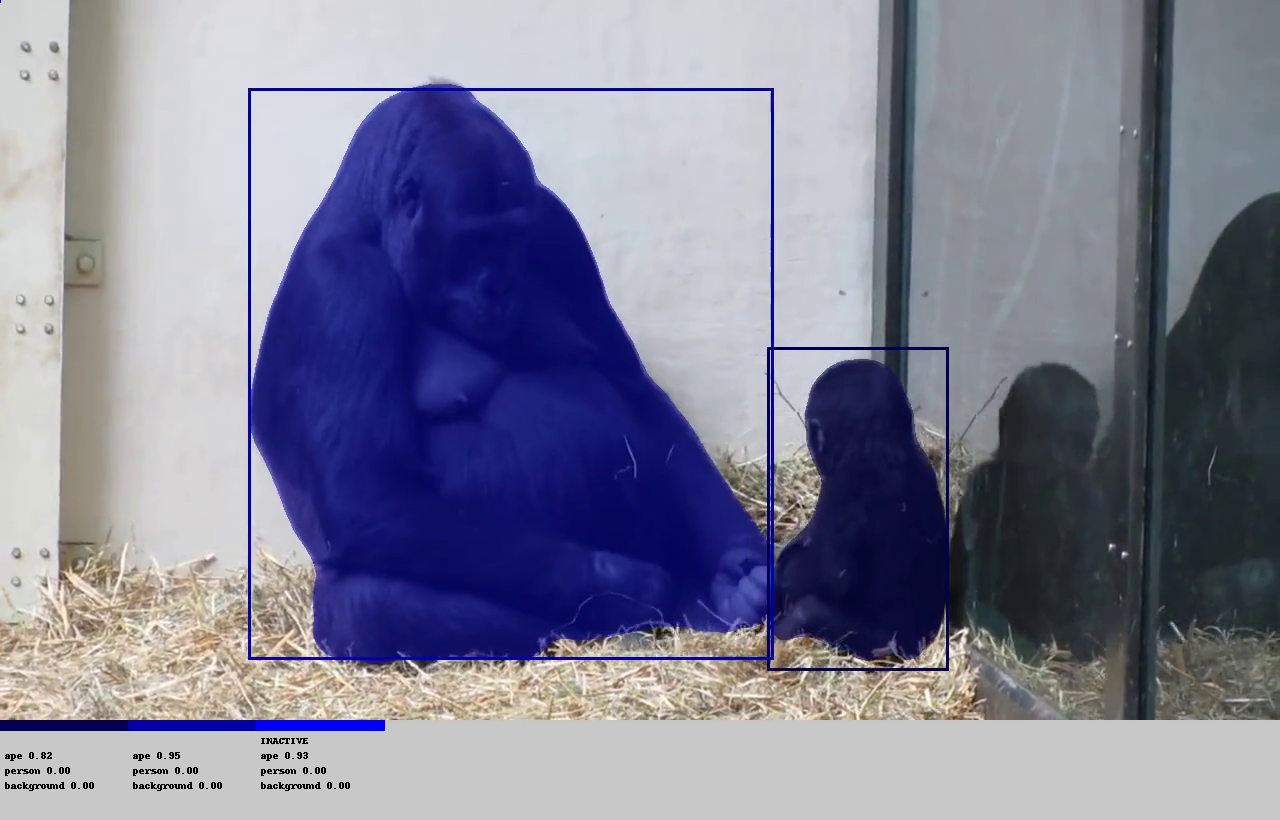}%
 \includegraphics[width=0.19\textwidth,trim={0.0cm 0.5cm 0.0cm 0.0cm}, clip]{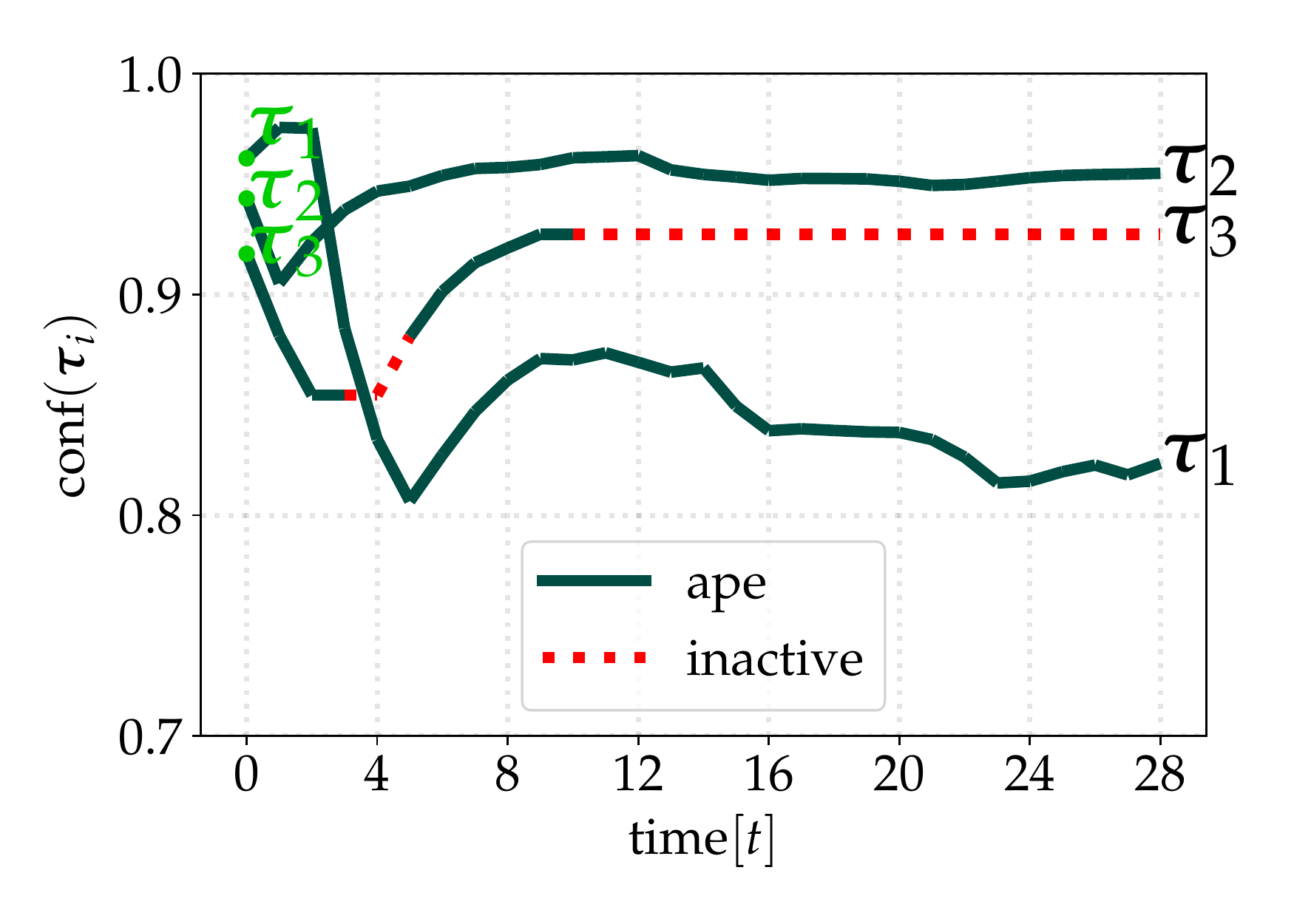}

 \caption{Here we provide an additional result comparing our approach (top row) with two configurations of our ablation study: i) association from~\cite{yang2019video} (middle row) and ii) scoring from~\cite{yang2019video} (bottom row). Frames are shown at times $t=\{3, 4, 7, 25\}$. The rightmost column shows predicted confidences and track identity. Clearly, noisy detections confuse the simple association method, initializing a lot of tracks. For the simple scoring (bottom row), the track $\tau_3$, a false positive covering half the ape on the left, is classified as an ape with high confidence. For this sequence, our approach (top row) is able to handle the noise and mark $\tau_3$ as background, resulting in a clear sequence of true positive tracks.}
 \label{fig:mirror_apes_compare}
\end{figure*}

\section{Additional Qualitative Results}
We supply additional qualitative results: a video showing our approach and a qualitative comparison for two of our ablation experiments.

\subsection{Qualitative Result Video}
We supply a video \texttt{qualitative\_results.mp4} with additional qualitative results. The video shows results produced by our final model with a ResNet101~\cite{he2016deep} backbone on the YouTubeVIS~\cite{yang2019video} validation set. The video depicts a variety of scenarios that our approach is able to handle. From single instances with fast motion, camera movements, to multiple similar instances. In the end we show some failure cases of our approach. For created tracks we show the top three class probabilities at the bottom of the video, in each of the sequences. Tracks deemed not to match any detection are marked as \emph{inactive}.



\subsection{Qualitative Ablation}
Section 4.1 of the main paper comprises an ablation study based on quantitative experiments. Here we highlight the difference between the proposed approach and two of the ablation configurations with a qualitative comparison. We compare the proposed approach with the \textbf{Association from \cite{yang2019video}} and the \textbf{Scoring from \cite{yang2019video}} configurations on a video of the YouTubeVIS validation set. The results are shown in Figure~\ref{fig:mirror_apes_compare}. In this sequnce, the association from \cite{yang2019video} leads to a high initialization rate of false positive tracks. The scoring from \cite{yang2019video} leads to a high score for a track that is a false positive. Our method in contrast has learnt to predict a low score when tracks are initialized, and later either reinforce or suppress that score. The false positive track in Figure~\ref{fig:mirror_apes_compare} is quickly suppressed and marked as background. Our approach suppresses noise in the detector and forms two accurate tracks.


\end{document}